\documentclass[11pt, onecolumn]{article}
\usepackage[top=1in, bottom=1in, left=1.25in, right=1.25in]{geometry}

\usepackage{amsfonts}
\usepackage{amsmath}
\usepackage{graphicx}
\usepackage{color, soul}
\usepackage{algorithm,algorithmic}
\usepackage{bm}
\usepackage{booktabs}
\usepackage{amssymb}
\usepackage{subfigure}
\usepackage{float}

\renewcommand{\arraystretch}{1.1}

\begin{document}

\title{Cross-label Suppression: A Discriminative and Fast Dictionary Learning with Group Regularization}
\date{April 24, 2017}

\author{Xiudong~Wang and Yuantao~Gu%
\thanks{The authors are with the Department of Electronic Engineering and Tsinghua National Laboratory for Information Science and Technology (TNList), Tsinghua University, Beijing 100084, P. R. China. The corresponding author of this work is Y. Gu (e-mail: gyt@tsinghua.edu.cn).
\newline
This paper has supplementary downloadable material available at http://gu.ee.tsinghua.edu.cn/publications/. The material includes MATLAB codes for the proposed algorithm and all experiments. Contact gyt@tsinghua.edu.cn for further questions about this work.
}
\vspace{2em}\\ 
Accepted for publication in \emph{IEEE Transactions on Image Processing}.
}

\maketitle

\begin{abstract}
This paper addresses image classification through learning a compact and discriminative dictionary efficiently.
Given a structured dictionary with each atom (columns in the dictionary matrix) related to some label, we propose cross-label suppression constraint to enlarge
the difference among representations for different classes.
Meanwhile, we introduce group regularization to enforce representations to preserve label properties of original samples, meaning the representations for the same class are encouraged to be similar.
Upon the cross-label suppression, we don't resort to frequently-used $\ell_0$-norm or $\ell_1$-norm for coding, and obtain computational efficiency
without losing the discriminative power for categorization.
Moreover, two simple classification schemes are also developed to take full advantage of the learnt dictionary.
Extensive experiments on six data sets including face recognition, object categorization, scene classification, texture recognition and sport action categorization are conducted, and the results show that the proposed approach can outperform lots of recently presented dictionary algorithms on both recognition accuracy and computational efficiency.

\end{abstract}

\section{Introduction}

Dictionary model has received much attention over the past few decades and has been widely adopted in a variety of applications, including image processing
\cite{K-SVD_M_2006}, \cite{doubleSparsity-2010}, clustering \cite{F_Wang_GroupSC2011},  and classification \cite{J_Wright_2009}, \cite{Mairal-2008}, \cite{J_Wang_LLC2010}.
The approach is built on the belief that a broad variety of signals such as images, video,
and audio can be well represented by a linear combination of a few elements from a set of representative patterns, where the whole set of representative patterns is called dictionary and its each element is called atom.
A dictionary acts as an effective tool for the sparse representation of a certain signal,
and the parsimony supplies a more meaningful way to capture the high-level semantics hidden in the signal \cite{F_Wang_GroupSC2011}, \cite{J_Wright_2009}, \cite{Mairal-2008}.
As a consequence, how to acquire a proper dictionary is crucial to the success of the sparse representation-based algorithms.
It has been shown that learning a task-specific dictionary from the training samples instead of off-the-shelf ones like various wavelets and Fourier bases can bring out superior results \cite{K-SVD_M_2006}, \cite{Olshausen-1997}.

Without using the label information of training samples, many algorithms have been presented mainly through sparse reconstruction for original signals and they can be referred to as unsupervised dictionary learning,
including the method of optimal directions (MOD) \cite{K_Engan_1999_MOD}, the classical K-SVD algorithm \cite{K-SVD_M_2006}, the least squares optimization \cite{H_Lee_2006}, and a structured dictionary model \cite{doubleSparsity-2010}.
In \cite{K_Engan_1999_MOD}, the dictionary is updated as a whole along the optimal direction efficiently.
In \cite{K-SVD_M_2006}, the dictionary is updated atom by atom and singular value decomposition is well combined.
A Lagrange dual is posted to learn the dictionary efficiently with fewer optimization variables than the primal in \cite{H_Lee_2006}.
Besides, in order to obtain low computational complexity, a sparse dictionary model \cite{doubleSparsity-2010} is proposed, in which each atom is required to be sparse over a known base dictionary.
Owing to this type of dictionary learning algorithms faithfully represent training signals, they are well adapted to reconstruction tasks, including image denoising and inpainting \cite{K-SVD_M_2006}, \cite{doubleSparsity-2010}.
However, it's not very advantageous to apply them for classification tasks due to the dictionary is learnt without exploiting the label information of the training samples.

A simple supervised dictionary can be directly obtained from the training samples without any refining process.
The sparse representation based classification (SRC) method is proposed for face recognition in \cite{J_Wright_2009}, and it directly adopts all training samples as the dictionary to represent signals.
For classification, the method classifies the query signal through identifying by which class training samples it can be best reconstructed.
Competitive results for face recognition are exhibited and robustness against noise is also demonstrated in the pioneering work.
However, the simple dictionary isn't compact and the discrimination power is not full exploited for recognition due to its lack of any further learning for more representative patterns.

In order to obtain a compact and discriminative dictionary for recognition,
a relatively small but refined dictionary is constructed by an iterative supervised learning with the labels of the training set in algorithms
\cite{Mairal-2008, Mairal-SDL2009, DKSVD_2010, DLSI_I_Ramirez_2010, COPAR-conf2012}, etc.
In these algorithms, both reconstruction errors and discriminative representations for classification are considered at the same time.
They exhibit impressive results in extensive recognition tasks, such as handwritten digit classification \cite{Mairal-2008, DLSI_I_Ramirez_2010, COPAR2014, FDDL_D_Yang2011},
face recognition \cite{ DKSVD_2010, COPAR2014, LCKSVD2013, M_Harandi_2015},
texture classification \cite{Mairal-2008, M_Harandi_2015, Y_Quan_MultiClassifier2016},
scene categorization \cite{COPAR2014, LCKSVD2013}, object classification \cite{COPAR2014, LCKSVD2013, D_Pham_2008}, and so on.

Nevertheless, the supervised algorithms for categorization mainly adopt $\ell_1$-norm \cite{J_Wright_2009, Mairal-2008, DLSI_I_Ramirez_2010, COPAR2014}, etc.,
or $\ell_0$-norm \cite{DKSVD_2010, LCKSVD2013, Y_Quan_MultiClassifier2016,D_Pham_2008} for pursuing sparse coding, which are generally optimized in an iterative manner owing to the nonsmoothness.
As a result, they are very computationally expensive for learning the dictionary and coding for classification.
For fast sparse coding, improved algorithms are presented \cite{KG-FastSparse2010} \cite{SKong-SADL2016}. \cite{KG-FastSparse2010} learns a non-linear, feed-forward predictor to produce the best possible approximation
of the sparse code. In addition, \cite{SKong-SADL2016} models sparse coding using the least-square solution with a shrinkage function, given a under-complete dictionary.

In this paper, we propose a novel supervised learning algorithm for a both discriminative and computationally efficient dictionary oriented to categorization.
We adopt a structured dictionary composed of label-particular atoms corresponding to some class and shared atoms commonly used by all classes.
Considering a signal should be mainly constructed by its closely associated atoms, i.e. the atoms with the same label as its and the shared ones,
we propose the cross-label suppression to constrain large coefficient appearance at other label-particular atoms rather than its closely associated ones.
The constraint enlarges difference among representations for diverse classes in terms of large coefficient positions.
Furthermore, we introduce the group regularization to improve the similarity of the representations for the same class and promote the label consistency.
Without using $\ell_0$-norm and $\ell_1$-norm for coding regularization, we employ a simpler coding regularization based on the three following reasons.
Firstly, even if we don't adopt $\ell_0$-norm or $\ell_1$-norm for coding regularization, the representation of one signal is approximately sparse,
due to the cross-label suppression enforce it to have a few large coefficients with lots of small ones which are not necessarily zero in the training stage. Moreover, these large coefficients are encouraged to mainly gather at the closely associated atoms of the signal,
and discriminative block structures are then possessed by the representations for diverse classes.
As a result, the discriminative power for classification is obtained, and it becomes of little significance to still make the intra-block components of one representation to be sparse.
Finally, an analytical solution can be obtained if $\ell_2$-norm is applied for coding, thus fairly accelerating the learning process and classification.

\begin{figure*}[t]
\begin{center}
\includegraphics[width=15cm]{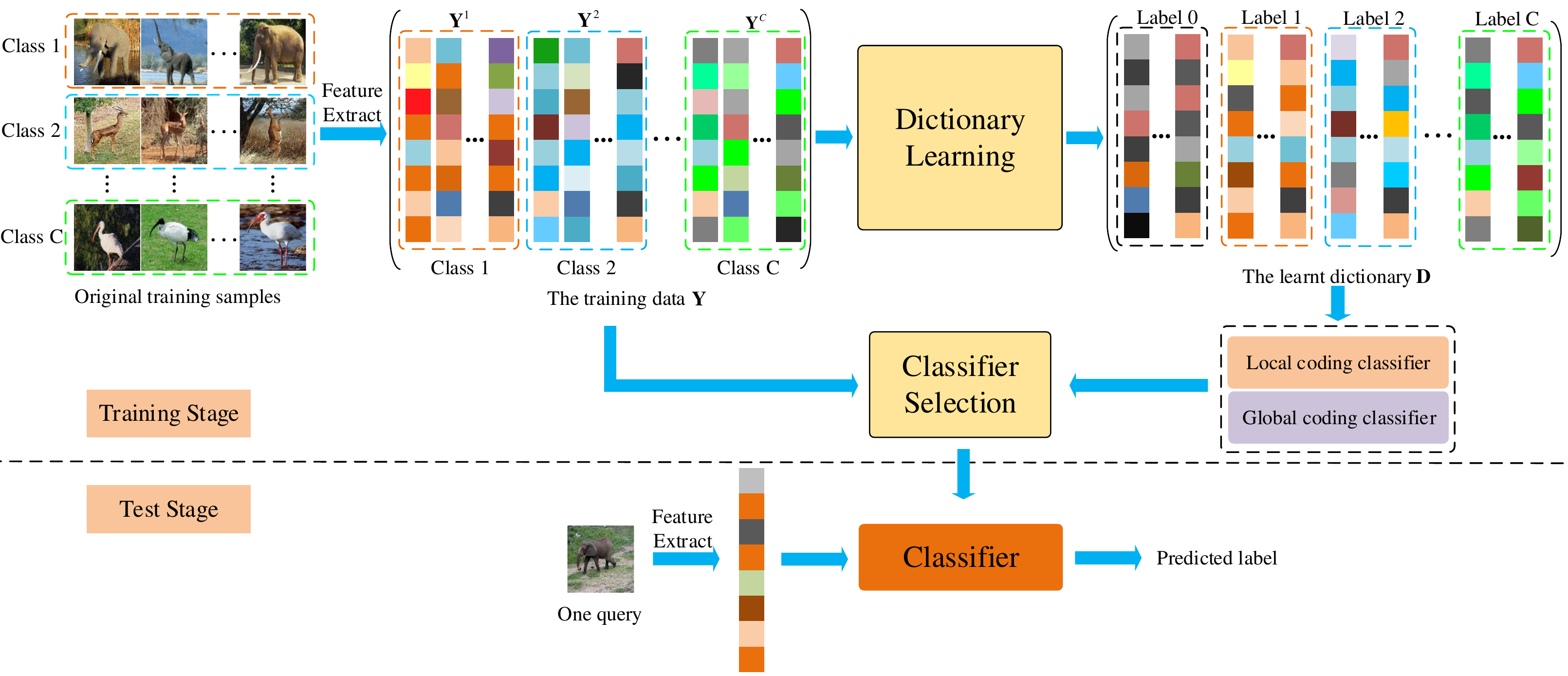}
\caption{The flowchart of our proposed algorithm for classification. The training stage mainly includes feature extraction, dictionary learning and classifier selection. The test stage mainly includes feature extraction and classification with the selected classifier.}
\label{flowchart}
\end{center}
\vspace{-2em}
\end{figure*}

According the dictionary design, two simple but practical classification schemes are developed in our work.
On account of that for the representation of one signal over the designed dictionary, large coefficients mainly occur on the atoms with the same label as its and the shared ones,
one classification scheme is accordingly proposed.
On the other hand, for one sample, owing to other label-particular atoms contribute little to its reconstruction, the signal should also be reconstructed well if only using its closely related ones.
As a result, another classification scheme is also developed.
In the case of classification, the better classification scheme can be easily selected for one specific dataset. The flowchart employed by our approach for classification is
shown in Figure \ref{flowchart}.

The main contributions in this paper are summarized as follows.
\begin{itemize}
\item
We adopt a structured dictionary consisting of label-particular atoms corresponding to some class and shared atoms used by all classes, and propose a novel label constraint called cross-label suppression to leverage the dissimilarities among representations for different classes.
\item
As for signals from the same class, the group regularization is proposed to preserve the label property and improve the similarity among their representations over the learnt dictionary.
\item
Upon the cross-label suppression as well as the signal reconstruction constraint, we don't employ widely-used $\ell_0$-norm or $\ell_1$-norm but a simpler $\ell_2$-norm for coding regularization, and computational efficiency is then gained without losing the discriminative power for classification.
\item
Finally, two simple classifiers are developed to cooperate with the learnt dictionary for image recognition and they can often bring out promising results.
\end{itemize}

The rest of this paper is organized as follows: In Section II and III, we review the related work and briefly introduce the preliminary on dictionary learning and the graph Laplacian, respectively.
In Section IV, we describe our cross-label suppression dictionary learning approach with the group regularization in details, including the formulation, optimization, classifiers and initialization.
Significantly, we conduct extensive experiments to evaluate the proposed algorithm in Section V and conclude our work in Section VI.

\section{Related work}
\subsection{Supervised Dictionary Learning}
In brief, the supervised dictionary learning algorithms for pattern recognition can be classified into three main categories.

The first category of developed dictionary learning algorithms learns a universal dictionary for all classes and imposes discriminative terms in the objective function to improve classification performance,
including \cite{Mairal-SDL2009}, \cite{DKSVD_2010}, \cite{LCKSVD2013}, \cite{D_Pham_2008}, \cite{Huang-2007}, \cite{Boureau-2010}, \cite{Lian-2010},  \cite{Yang-2010}, \cite{W-Wang-2016}.
Specifically, Fisher criterion for enhancement is employed in \cite{Huang-2007}, and softmax discriminative term is incorporated into the cost function by \cite{Mairal-SDL2009}, \cite{Boureau-2010}.
Additionally, a classifier is introduced for joint learning with the dictionary during training in \cite{D_Pham_2008}, \cite{Mairal-SDL2009}, \cite{DKSVD_2010}, \cite{LCKSVD2013}, \cite{Lian-2010}, \cite{Yang-2010},
where hinge loss function \cite{Lian-2010}, \cite{Yang-2010}, logistic loss function \cite{Mairal-SDL2009},
and linear prediction cost \cite{DKSVD_2010}, \cite{LCKSVD2013}, \cite{D_Pham_2008}, are adopted for training the classifier, respectively.
Upon employing a linear classifier adopted in \cite{DKSVD_2010}, \cite{LCKSVD2013} additionally proposes the label consistency constraint in the objective function to leverage the discriminative power,
and achieves impressive results in multiple recognition tasks such as face recognition, object categorization, and sports action recognition.

The second strategy for promoting the discriminability learns kinds of structured dictionaries,
including a set of class-specific dictionaries \cite{Mairal-2008}, \cite{DLSI_I_Ramirez_2010}, \cite{JDL-2012}, \cite{CSDL2014},
one universal dictionary with each atom labeled like training signals \cite{LCKSVD2013},
and a set of class-specific dictionaries combined with a universal dictionary \cite{COPAR2014}, \cite{FDDL_D_Yang2011}.
\cite{Mairal-2008} introduces the softmax term among multiple class-specific dictionaries based on the K-SVD model \cite{K-SVD_M_2006}, and apply them for texture segmentation and scene analysis.
\cite{DLSI_I_Ramirez_2010} learns a class-specific dictionary for each class with sparse coding, and impose the mutual incoherence among these dictionaries,
attaining an excellent performance for digit and audio classification.
Upon on \cite{DLSI_I_Ramirez_2010}, \cite{CSDL2014} additionally introduces self-dictionary incoherence term for fine-grained image categorization.
Furthermore, inspired by the application of the shared sub-dictionary for clustering \cite{F_Wang_GroupSC2011}, \cite{COPAR2014} employs a common sub-dictionary shared by all the classes other than class-specific dictionaries for classification.
This strategy is also used in \cite{JDL-2012}, \cite{CSDL2014}.
\cite{FDDL_D_Yang2011} and  \cite{COPAR2014} employ a joint strategy of learning a global dictionary and class-specific dictionaries at the same time,
expecting both the global dictionary and each class-specific dictionary possess a good reconstruction for the corresponding class samples.

Different from the above two categories of supervised dictionary learning,
the third type of learning a discriminative dictionary assumes all the samples correspond to another space with different dimension from the original one,
including kernel-based methods \cite{M_Harandi_2015}, \cite{kernel-SR-2013}, \cite{JointKernel-2015}, \cite{RobustKernel-2015}, \cite{MultipleKenel-2014}, \cite{Riemainn-manifold2}
and manifold-based algorithms \cite{M_Harandi_2015}, \cite{Riemainn-manifold2}, \cite{Riemainn-manifold1}.
Instead of the direct linear construction in the original space, these algorithms firstly need to map both signals and atoms into another space and then conduct linear constructions for signals with the dictionary,
which are often used to address nonlinear problems.
In kernel-based dictionary learning, multiple kernels have been jointly employed for better results in \cite{MultipleKenel-2014},
unlike \cite{kernel-SR-2013}, \cite{JointKernel-2015}, \cite{RobustKernel-2015} with just one single kernel.
Besides, Riemannian manifolds are applied in \cite{Riemainn-manifold2}, \cite{Riemainn-manifold1} and Grassmann manifolds are employed in \cite{M_Harandi_2015}.

To make representations discriminative, we employ a structured dictionary in a more flexible way.
Explicitly, we propose the cross-label suppression to constrain large coefficient appearance at other label-particular atoms rather than its closely associated ones.
Unlike multiple class-specific dictionaries-based approaches such as \cite{Mairal-2008,DLSI_I_Ramirez_2010, JDL-2012, COPAR2014},
the label constraint don't fully cut off the collaboration among atoms with different labels for reconstructing samples during the learning process.
Besides, we don't need to predefine discriminative sparse codes to utilize the dictionary structure like \cite{LCKSVD2013}.
In \cite{LCKSVD2013}, owing to all the nonzero coefficients in the predefined discriminative sparse codes for each class are identically set to 1,
nonzero coefficients of one learnt sparse code are forced to be equal to some extent, and it isn't very convincing.

\subsection{Related work on the graph Laplacian}

The graph Laplacian as a very flexible tool for representing and processing signals is applied in many domains, including dimensionality reduction \cite{graph-lapReduceDim-2003},
classification and clustering \cite{spectral-clustering-2002}, \cite{local-global-consistency-2003}, \cite{graph-SC-2011}, \cite{supGraphDL-arxiv2016},  and image smoothing \cite{graph-heatKernel-2008}.
\cite{graph-lapReduceDim-2003} exploits the geometry structure incorporating neighborhood information of the data set and
proposes Laplacian eigenmaps for dimensionality reduction and data representation, which possess locality-preserving properties.
Based on the $k$ ($k\in\mathbb{N}$) largest eigenvectors of a normalized Laplacian, \cite{spectral-clustering-2002} proposes a classical spectrum-based approach for clustering.
In semi-supervised learning, \cite{local-global-consistency-2003} imposes a smoothness constraint on the classifying function through the Laplacian of the intrinsic structure revealed by known labeled and unlabeled data points,
and attain encouraging results for handwritten digit recognition and text classification.
\cite{graph-SC-2011} presents graph regularized sparse coding with respect to a unsupervised dictionary for image presentation using the Laplacian as a smooth operator, and validate its effectiveness on both classification and clustering.
\cite{supGraphDL-arxiv2016} introduces two adaptive Laplacians for dictionary learning and sparse coding, respectively, and apply them to the single label recognition and multi-label classification.
Considering the image intensity diffusion,
\cite{graph-heatKernel-2008} accomplishes the image smoothing by convolving original images with the heat kernel governed by the Laplacian of the graph, which is constructed by pixel lattices.

There is a significant and common principle applied in \cite{graph-lapReduceDim-2003}, \cite{spectral-clustering-2002},
\cite{local-global-consistency-2003}, \cite{graph-SC-2011} that if two data points are close,
their corresponding maps (e.g. reduced representations \cite{graph-lapReduceDim-2003} and classification vectors \cite{local-global-consistency-2003} ) should be also close to each other.
The principle can guarantee the maps preserve locality properties and can reflect the geometrical structure revealed by original data points, resulting in better performance.

In our work, we also incorporate the Laplacian tool into our supervised dictionary learning model.
Differently, the graph is constructed according to label information of samples rather than $k$ nearest neighbors ($k \in \mathbb{N}$) criterion \cite{graph-lapReduceDim-2003}, \cite{graph-SC-2011}, or
 $\epsilon$-neighborhoods ($\epsilon>0$) criterion \cite{graph-lapReduceDim-2003}, \cite{spectral-clustering-2002}, \cite{local-global-consistency-2003}, \cite{graph-heatKernel-2008}.
 Although this direct construction method is relatively easy without any threshold such as $k$ and $\epsilon$, the efficacy is well demonstrated in our experiments.

\section{Background}

In this section, we will review the general dictionary learning model and the Laplacian briefly.

\subsection{General dictionary learning}
Let ${\bf D}=[{\bf d}_1,\dots,{\bf d}_K]\ \in{\mathbb{R}^{M\times K}}$ denote the dictionary to be learnt, where $M$ and $K$ signify the dimensionality of each atom and the total number of atoms in the dictionary, respectively.
Considering samples from $C$ classes ($C \in \mathbb{N}$), let ${\bf Y}^c=[{\bf y}_1^c,\dots,{\bf y}^c_{N^c}]\ \in{\mathbb{R}^{M\times N^c}}$ denote the $c^{th}$ class training samples with each sample ${\bf y}^c_i \in \mathbb{R}^{M\times 1}$ ($i=1,\dots,N^c$,
where the sample number is $N^c$), and let ${\bf X}^c=[{\bf x}^c_1,\dots,{\bf x}^c_{N^c}]\ \in \mathbb{R}^{K\times N^c}$ denote the corresponding representations of ${\bf Y}^c$ over the dictionary, $n=1,\dots,C$.
The classical dictionary learning model can be formulated as
\begin{align}
\min_{\bf D,X} \sum_{c=1}^{C}\left({\left\|{\bf Y}^c-{\bf D}{\bf X}^c\right\|_F^2+\beta\sum_{j=1}^{N^c}\|{\bf x}^c_j\|_p}\right)
\qquad \text{s.t.}\quad \|{\bf d}_k\|=1, \forall k,
 \label{dictionary_learning}
\end{align}
where ${\bf X}= [{\bf X}^1,\dots,{\bf X}^C] \in \mathbb{R}^{K\times N}$ ($N=\sum_{c=1}^{C}N^c$, the total number of all the training samples) denotes the representations of all the training samples
${\bf Y}=[{\bf Y}^1,\dots,{\bf Y}^C] \in \mathbb{R}^{M\times N}$, and $\ell_p$-norm is used for sparse regularization with $p$ frequently set as $1\ \text{and} \ 0$ like SRC \cite{J_Wright_2009} and K-SVD \cite{K-SVD_M_2006}.
To avoid trivial solutions, it's common to constrain the energy of each atom to be normalized.
In summary, the first term represents the reconstruction error of ${\bf Y}^c$ by ${\bf X}^c$ and the dictionary ${\bf D}$, and the second term denotes the sparse regularization with a scalar $\beta$.

Besides, it's worth mentioning that when $\ell_0$-norm is adopted, the sparse regularization is often substituted with $\|{\bf x}_i^c\|_0\le T$ ($\forall i, c$), in which $T$ is the sparsity constraint factor.
Then the model can be reformulated as
\begin{align}
\min_{\bf D,X} \sum_{c=1}^{C}{\left\|{\bf Y}^c-{\bf D}{\bf X}^c\right\|_F^2} \qquad\text{s.t.}\; \quad \|{\bf x}_i^c\|_0\le T \;{\rm and}\;  \|{\bf d}_k\|=1, \forall i,c, k.
 \label{DL-L0}
\end{align}

Thus, minimizing above objective functions intends to achieve that each sample in ${\bf Y}$ can be well represented as a sparse linear combination of those atoms in the learnt dictionary.
\subsection{The graph Laplacian}
The Laplacian matrix \cite{chung_spectral_1997} of an $N$-vertex undirected graph $\mathcal{G}$ with the vertex set $\mathcal {V}$ is defined as
$$
{\bf L=M-W},
$$
where ${\bf W}$ is the adjacency matrix of $\mathcal{G}$, and ${\bf M}$ is the degree matrix, which is a diagonal matrix diag$\{m_1,\cdots,m_N\}$ with $m_i=\sum_{j}w_{ij}$.
Since $w_{ij}=w_{ji}$ for undirected $\mathcal{G}$, ${\bf L}$ is symmetrical and actually positive semi-definite. It can be normalized as follows
\begin{align}
\tilde{\bf L}&={\bf M}^{-\frac{1}{2}}{\bf L}{\bf M}^{-\frac{1}{2}} \nonumber\\
             &={\bf I}-{\bf M}^{-\frac{1}{2}}{\bf W}{\bf M}^{-\frac{1}{2}},
\end{align}
where ${\bf I}$ denotes the identity matrix.
Given a map ${\bf f}=[f_1,\dots,f_N]^{\rm T}\in \mathbb{R}^N$ mapping the graph to a line like the case in \cite{graph-lapReduceDim-2003}, in which $f_i$ corresponds to the vertex $i$,
its variation among neighboring vertices can be attained as
\begin{align}
 {\text {Var}}({\bf f}) &= \frac{1}{2}\sum_{i\sim j}w_{ij}\left(\frac{f_i}{\sqrt{m_i}}-\frac{f_j}{\sqrt{m_j}}\right)^2 \nonumber\\
                        &= {\bf f}^\text{T}\tilde{\bf L}{\bf f}.  \nonumber
\end{align}
The variation indicates the smoothness of the map $\bf{f}$ among neighboring vertices. Specifically the smaller its variation is, the more smooth the map is.

\section{cross-label suppression dictionary learning with group regularization}

We intend to leverage the supervised information of training samples to make the learnt dictionary more discriminative for classification, in addition to being constructive.
For the end, the cross-label suppression and group regularization are proposed.
Inspired by \cite{F_Wang_GroupSC2011, COPAR2014},
we jointly consider shared atoms for all classes and label-particular atoms mainly associated with some particular class in our algorithm.

\subsection{Object function}
For designing a compact and discriminative dictionary, each atom should be representative and have a particular semantic meaning.
As a result, each atom in the dictionary can be associated with a particular label, corresponding to some specific class.
Nevertheless, there may be lots of common features among various classes such as similar background, and shared atoms should be taken into account also, corresponding to all the classes.
Similarly, the shared atoms is linked to the common label (denoted as label 0 here) rather than any particular one.
Then, similar to training samples, each atom in the dictionary has its own label.
Denote $\mathcal{I}^0$ as the index set for shared atoms, $\mathcal{I}^i$ as the index set for atoms with the $i^{th}$ class label in the dictionary, $i=1, \dots, C$,
and $\mathcal{I}$ as the index set of all the atoms, where $C$ signifies the total number of classes.
To obtain more powerful discrimination to distinguish different classes, it is desirable that for signals from the same class,
representation coefficients over the dictionary will intensively locate on its own particular atoms and the shared ones.
Therefore, representation coefficients locating out of their corresponding particular atoms and the shared ones need to be suppressed to some extent.

Besides, to describe the geometry structure of the labels for all the training samples, we can construct a graph for them.
In the light of label information,
provided one sample corresponds to one vertex, the vertices related to the same class samples are neighboring to each other and connected,
while the vertices for different classes are far from each other and aren't connected.
Each class corresponds to a densely connected subgraph with each two vertices connected, forming a group.
There are in total $C$ groups consisting of $C$ complete subgraphs, while there are no connections among those groups.

\begin{figure}[t]
\begin{center}
\includegraphics[width=8.5cm]{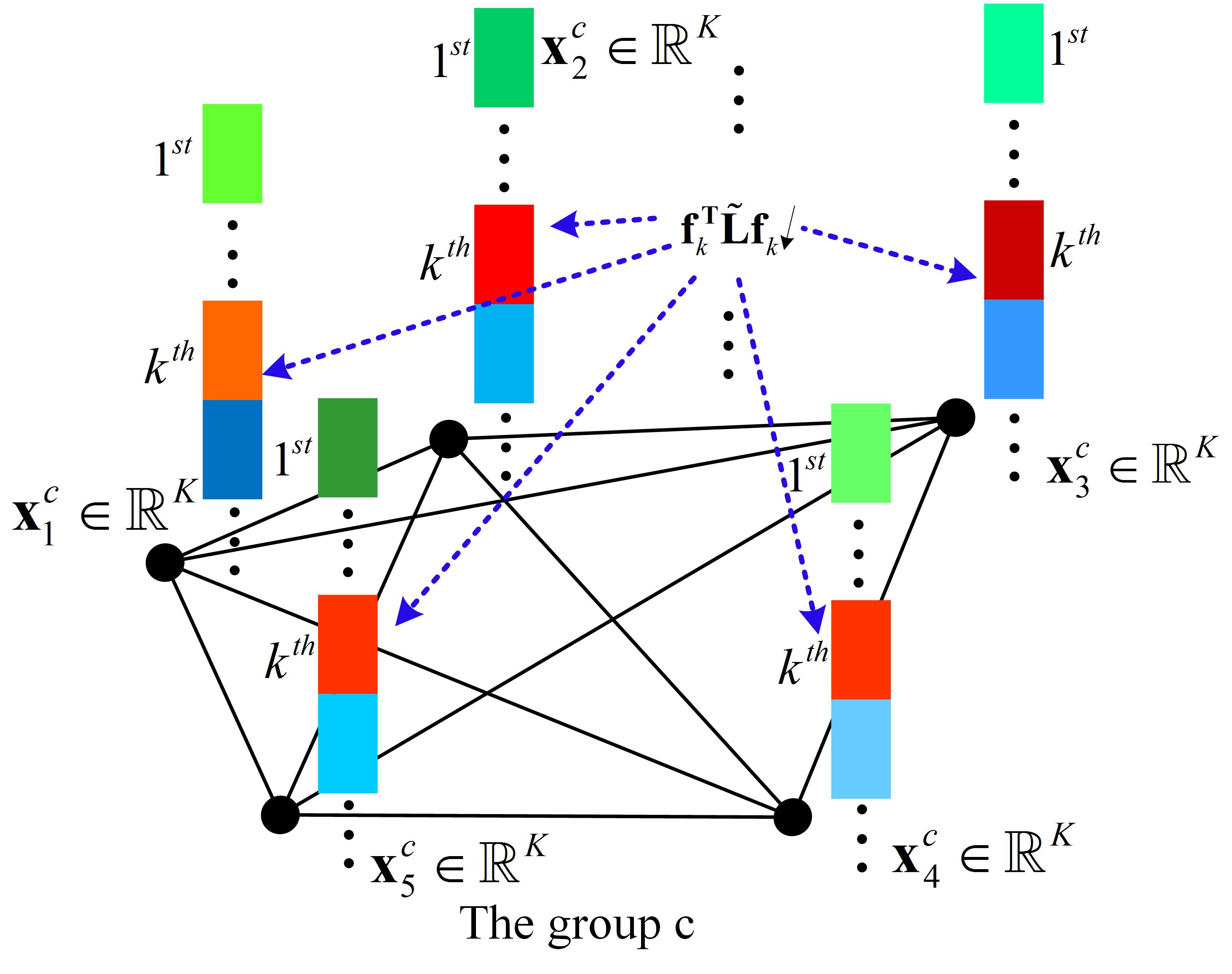}
\caption{Illustration for the effect of the group regularization on the representations for the class c samples.
Each vertex corresponds to one sample and then relates to its representation. Each component of the representation is denoted as one colored block with similar colors signifying
the values are similar. Keeping the map variation ${\bf f}_k^{\rm T}\tilde{\bf L}{\bf f}_k$ small will force the the $k^{th}$ components of the representations to be similar.
}
\label{lapwork}
\end{center}
\vspace{-0.5em}
\end{figure}

Considering the training samples ${\bf Y}=[{\bf Y}^1,\cdots,{\bf Y}^C] \in \mathbb{R}^{M \times N}$ and their corresponding representations ${\bf X}=[{\bf X}^1,\cdots,{\bf X}^C] \in \mathbb{R}^{K \times N}$,
we can define $K$ graph maps with mapping the graph to a line consisting of $N$ points as
$$
	{\bf f}_k = \left[\begin{array}{c}{\bf f}_k^1\\{\bf f}_k^2\\ \vdots\\ {\bf f}_k^C\end{array}\right]\in \mathbb{R}^{N\times 1},\quad \forall k=1,2,\dots,K,
$$
where
$$
{\bf f}_k^c = \left[{\bf x}_1^c(k),{\bf x}_2^c(k),\dots,{\bf x}_{N^c}^c(k) \right]^{\rm T} \in \mathbb{R}^{N^c},\quad  c = 1,\dots,C.
$$
Likewise ${\bf Y}^c$, ${\bf X}^c$, ${\bf x}^c_j$, $N^c$ ($c=1,\dots,C$), $M$, $N$, and $K$ have the same meanings as those described for  (\ref{dictionary_learning}) and ${\bf x}^c_j(k)$ denotes the $k^{th}$
component of ${\bf x}^c_j$, which also corresponds to the $k^{th}$ atom in the dictionary.
Therefore, the total variation for these $K$ maps can be attained as
\begin{align}
{\rm totalVar} &= \sum_{k=1}^{K} {\bf f}_k^{\rm T}\tilde{\bf L}{\bf f}_k  \nonumber\\
               &={\rm tr}\left( \left[ \begin{array}{c} {\bf f}_1^{\rm T}\\ \vdots\\ {\bf f}_K^{\rm T} \end{array} \right]\tilde{\bf L}\ [{\bf f}_1,\cdots,{\bf f}_K] \right) \nonumber \\
               &={\rm tr}( {\bf X}\tilde{\bf L}{\bf X}^{\rm T} ),
\end{align}
where the normalized Laplacian matrix of the whole graph is assumed as $\tilde{\bf L}\in \mathbb{R}^{N\times N}$, and ${\rm tr}(\cdot)$ denotes the trace operator for matrices.
Keeping the $k^{th}$ map variation ${\bf f}_k^{\rm T}\tilde{\bf L}{\bf f}_k$ ($k=1,\,\dots,\,K$) small will force the $k^{th}$ components to be similar at neighboring vertices, as illustrated in Figure \ref{lapwork}.
To well preserve the label property, the representations for the same class samples should be similar and then the total variation ought to be kept small to some extent.

\begin{figure}[t]
\begin{center}
\includegraphics[width=8.5cm]{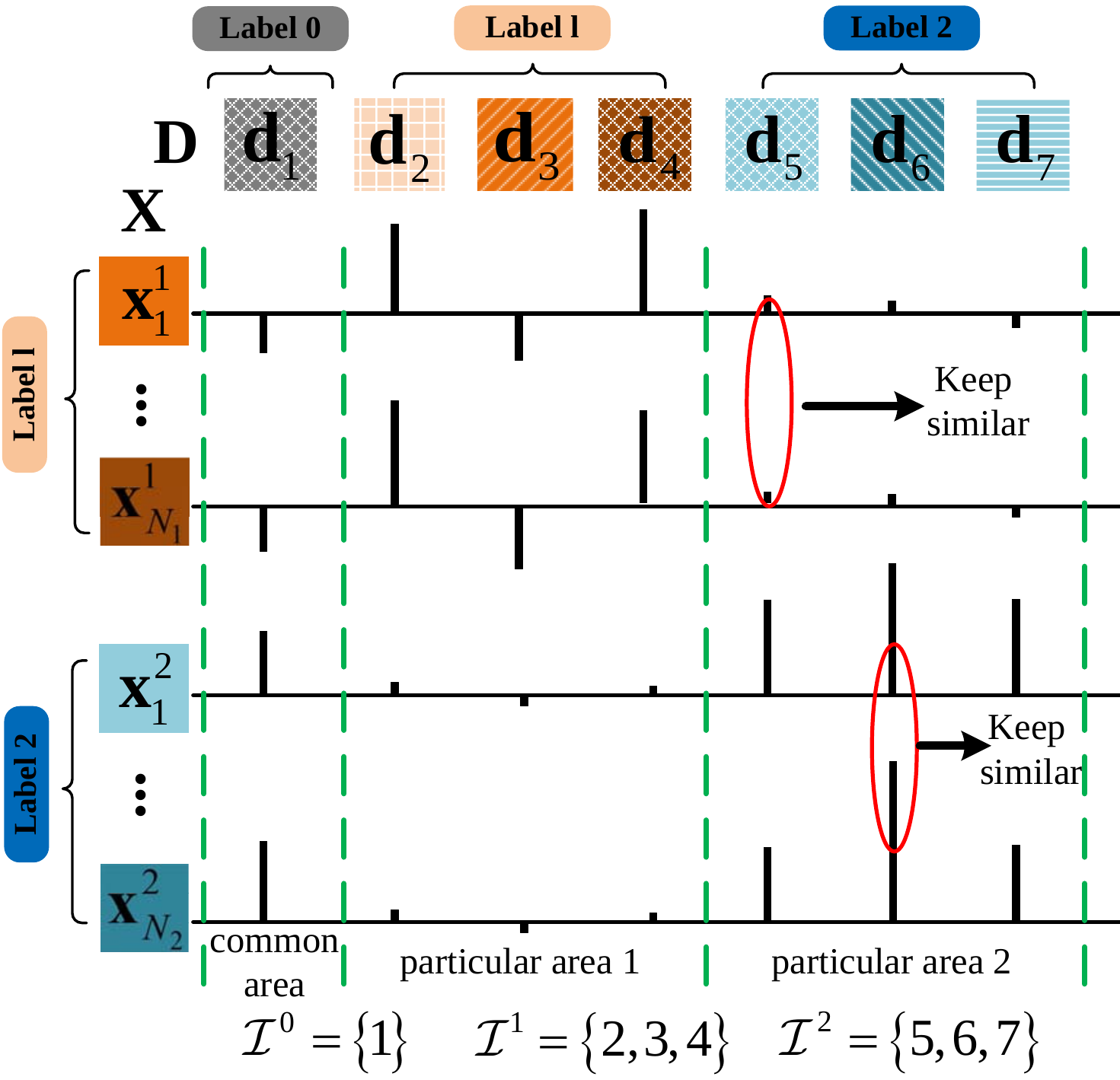}
\caption{Scheme illustration of cross-label suppression dictionary learning with group regularization.
Given samples from different classes,
we expect their representations over the dictionary are remarkably distinct with large coefficients locating in different areas.
However, for samples from the same class,
we hope their representations to preserve the label property and be similar.}
\label{whole-model}
\end{center}
\end{figure}

As illustrated in Figure \ref{whole-model}, we propose the following dictionary learning model
\begin{align}
&\min_{\bf D,X} \sum_{c=1}^C \bigg\{ \left\|{\bf Y}^c-{\bf D}{\bf X}^c\right\|_F^2 + \beta \sum_{j=1}^{N^c}\left\|{\bf x}^c_j\right\|_2^2 \bigg\} 
+ \lambda \sum_{c=1}^{C} \left\| {\bf P}^c{\bf X}^c\right\|_F^2 +\gamma \text {tr}({\bf X}\tilde{\bf L}{{\bf X}}^{\rm T})\nonumber \\
&\text{s.t.}\ \|{\bf d}_k\|_2=1, \forall k,
\label{costv1}
\end{align}
where in addition to the first term denoting the reconstruction error and the second term signifying the representation regularization,
the two later terms represent the cross-label suppression and the group regularization, respectively.
Beside ${\beta}$ described in  (\ref{dictionary_learning}), $\lambda$ and ${\gamma}$ are the scalars controlling the relative contribution for the corresponding items.

In the cross-label suppression term $\|{\bf P}^c{\bf X}^c\|_F^2$,  ${\bf P}^c\in \mathbb{R}^{K\times K}$ denotes
the extracting matrix for picking up coefficients from the representation of one signal,
which locate at other label-particular atoms rather than its closely associated atoms, defined as
\begin{align}
	{\bf P}^c(m,n) = \left\{\begin{array}{ll}
	1, & m = n \; {\rm and} \; m\in \mathcal{I}\backslash(\mathcal{I}^0 \cup \mathcal{I}^c);\\
	0, & {\rm elsewhere},
	\end{array}\right.
\end{align}
where $c=1,\dots,C$ and ${\bf P}^c(m,n)$ denotes the $(m,n)^{th}$ entry of ${\bf P}^c$.
Then
$$
	({\bf P}^c{\bf x}^c_i)(m) = \left\{\begin{array}{ll}
	{\bf x}_i(m),& m\in \mathcal{I}\backslash(\mathcal{I}^0 \cup \mathcal{I}^c);\\
	0, & {\rm elsewhere},
	\end{array}\right.
$$
where $({\bf P}^c{\bf x}^c_i)(m)$ and ${\bf x}_i(m)$ denote the $m^{th}$ components of the corresponding vectors, respectively.
Thus, for the representation of some signal, the constraint suppresses large coefficient occurrence at other label-particular atoms to some extent,
and encourages large coefficients to mainly gather towards its closely related atoms.
Furthermore, that constraint with a proper scalar can bring the representation with only a few large coefficients and lots of small ones in the training stage, that is,
an approximately sparse representation.

Due to the cross-label suppression works, we don't employ traditional sparsity-induced norm such as ${\ell_0}$-norm, and ${\ell_1}$-norm for regularizing representations.
Furthermore, analytical solution is then obtained for the representation unlike the iterative optimization for the representation with $\ell_1$-norm or $\ell_0$-norm regularization.
As a result, it will become more efficient for coding over the dictionary.

As for the group regularization, owing to the whole graph consists of $C$ groups isolated from each other, the normalized Laplacian matrix can be formed as
\begin{align}
\tilde{\bf L}=\left[
\begin{matrix}
         \tilde{\bf L}^1 &           &        &\\
                   & \tilde{\bf L}^2 &        &\\
                   &           &\ddots  &\\
                   &           &        &\tilde{\bf L}^C
\end{matrix}
\right] \in \mathbb{R}^{N\times N}
\end{align}
in which $\tilde{\bf L}^c \in\mathbb{R}^{N^c\times N^c}$ ($c=1,\ 2,\ \cdots,\ C$) denotes the normalized Laplacian of the $c^{th}$ class corresponding subgraph.
Considering $N^c$ training samples for the $c$ class and each two vertices are connected, $\tilde{\bf L}^c$ can be derived as
\begin{align}\label{eq:definetildeLc}
\tilde{\bf L}^c =\frac{1}{N^c-1} \left[
 \begin{array}{cccc} N^c-1   & -1      & \cdots  & -1     \\
                      -1     & N^c-1   & \cdots  & -1     \\
                      \vdots & \vdots  & \vdots  & \vdots \\
                      -1     & -1      & \cdots & N^c-1
 \end{array} \right].
\end{align}

Based on $\tilde{\bf L}={\rm diag}(\tilde{\bf L}^1,\ \cdots,\ \tilde{\bf L}^C)$, the objective function in (\ref{costv1}) can be reformulated as follows
\begin{align}
&\min_{\bf D,X}\sum_{c=1}^C\bigg\{\left\|{\bf Y}^c-{\bf D}{\bf X}^c\right\|_F^2 + \beta \sum_{j=1}^{N^c}\left\|{\bf x}^c_j\right\|_2^2 
                          +\lambda \left\| {\bf P}^c{\bf X}^c\right\|_F^2 +\gamma \text {tr}\left({\bf X}^c\tilde{\bf L}^c({\bf X}^c)^ \text{T}\right)\bigg\}\nonumber \\
                           &  \text{s.t.}\ \|{\bf d}_k\|_2=1, \forall k.
\label{costv2}
\end{align}

In the following section, we mainly describe the optimization procedure for the proposed method.

\subsection{Optimization}

The objective function of the cross-label suppression dictionary learning problem (\ref{costv2}) is not convex.
However, we can alternatively update the representations and the dictionary in principle like those in \cite{K-SVD_M_2006}, \cite{H_Lee_2006},
while keeping all the remaining variables fixed.
Overall optimization process is shown in Algorithm 1.

\begin{table}[t]\nonumber
\renewcommand{\arraystretch}{1.2}
\begin{center}
\begin{tabular}{l}
\toprule
{\bf Algorithm 1:} Cross-label suppression dictionary learning with group \\
\hspace{5.8em}                 regularization \\
\hline
{\bf Input:}\hspace{0.3em}  ${\bf Y}=[{\bf Y}^1,\dots,{\bf Y}^C]$, $\beta, \gamma, \lambda$; \\
{\bf Initialization:}\hspace{0.3em}Initialize label-particular atoms using k-means with \\
                                   corresponding class training samples, and initialize shared ones \\
                                   using k-means with all the residual produced by using (\ref{initResi});\\
                                   Initialize representations by using (\ref{initCode}).\\
{\bf Repeat} \\
\hspace{1.3em} 1) Update representations while fixing the dictionary\\
\hspace{2.2em} for $c=0,\ 1,\ 2,\ \dots,\ C$\\
\hspace{2.8em} Update the ${c^{th}}$ class codes ${\bf X}^c$ by using (\ref{codeMatrix});\\
\hspace{1.3em} 2) Update the dictionary while fixing representations\\
\hspace{2.2em} for $c=0,\ 1,\ 2,\ \dots,\ C$\\
\hspace{2.8em} ${\bf Z}={\bf Y}- \sum_{ k\notin \mathcal{I}^c } {\bf d}_k \bar{\bf x }_k$\\
\hspace{2.8em} for $i\in\mathcal{I}^c$\\
\hspace{3.6em} Update the atom ${\bf d}_i$ by using  (\ref{Y-updateAtom}) and (\ref{atomNormalization});\\
\hspace{2.8em} end \\
\hspace{2.2em} end \\
{\bf Until:} The objective function converges or the maximum number\\
\hspace{2.8em}   of iterations is reached.\\
{\bf Output:} The dictionary ${\bf D}$.\\
\bottomrule
\end{tabular}
\end{center}
\vspace{-1em}
\end{table}

1) {\itshape Update representations:} For {the ${c^{th}}$ class representations ${\bf X}^c$, optimizing the part depending on it goes as
 \begin{align}
\min_{{\bf X}^c} \left\|{\bf Y}^c-{\bf D}{\bf X}^c\right\|_F^2+\beta\sum_{k=1}^{N^c}\left\|{\bf X}^c_k\right\|_2^2 +\lambda \left\| {\bf P}^c{\bf X}^c\right\|_F^2 
+\gamma \text{tr}\left({\bf X}^c\tilde{\bf L}^c{\left({\bf X}^c\right)}^\text{T}\right).
 \end{align}
The group regularization term can be unfolded as
\begin{align}
 \text{tr}\left({\bf X}^c\tilde{\bf L}^c{\left({\bf X}^c\right)}^{\rm T}\right)&=\text{tr}\left([{\bf x}^c_1\, {\bf x}^c_2\, \cdots\, {\bf x}^c_{N^c}]\tilde{\bf L}^c[{\bf x}^c_1\, {\bf x}^c_2\, \cdots \,{\bf x}^c_{N^c}]^{\rm T}\right)\nonumber\\
 &= \text{tr}\left(\sum_{i,j}\tilde{\bf L}^c(i,j)\,{\bf x}^c_i\, ({\bf x}^c_j)^{\rm T}\right) \nonumber\\
 &= \sum_{i,j}\tilde{\bf L}^c(i,j)\, \text{tr}\left(({\bf x}^c_j)^{\rm T}{\bf x}^c_i\right)\nonumber\\
 &= \sum_{i,j}\tilde{\bf L}^c(i,j)({\bf x}^c_j)^{\rm T}{\bf x}^c_i,
\end{align}
where $\tilde{\bf L}^c(i,j)$ denotes $(i,j)^{th}$ entry of $\tilde{\bf L}^c$.
We update ${\bf X}^c$ code by code, and then ${\bf x}^c_i$ is optimized by the following process,
 \begin{align}
\hat{\bf x}^c_i=\text{arg} \min_{{\bf x}^c_i} \left\|{\bf y}^c_i-{\bf D}{\bf x}^c_i\right\|_2^2+\beta\left\|{\bf x}^c_i\right\|_2^2 
 + \lambda\left\|{\bf P}^c{\bf x}^c_i\right\|_2^2 + \gamma  \sum_{k,j}\tilde{\bf L}^c(k,j)({\bf x}^c_j)^{\rm T}{\bf x}^c_k .\nonumber
 \end{align}

Then the solution is obtained
\begin{align}
 \hat{\bf x}^c_i = \left( {\bf D}^\text{T} {\bf D}+\lambda {\left({\bf P}^c\right)}^\text{T}{\bf P}^c +(\beta + \gamma \tilde{\bf L}^c(i,i)){\bf I}\right)^{-1}
                                  \left( {\bf D}^\text{T}{\bf y}^c_i-\gamma \sum_{j \not= i}\tilde{\bf L}^c(i,j){\bf x}^c_j \right),\label{codeVector}
 \end{align}
 where ${\bf I} \in \mathbb{R}^{K\times K}$ denotes an identity matrix. Denoting $\bar {\bf L}^c $  as $\tilde{\bf L}^c-{\bf I}^c$ with the identity matrix ${\bf I}^c\in\mathbb{R}^{N^c\times N^c}$ and using  \eqref{eq:definetildeLc},
 we can reform  (\ref{codeVector}) into a matrix version for ${\bf X}^c$ update,
 \begin{align}
 \hat{\bf X}^c = \left( {\bf D}^\text{T}{\bf D}+\lambda {\left({\bf P}^c\right)}^\text{T}{\bf P}^c +(\beta + \gamma){\bf I}\right)^{-1}\!\!\!\left({\bf D}^\text{T}{\bf Y}^c-\gamma {\bf X}^c\bar{\bf L}^c\right).
 \label{codeMatrix}
 \end{align}

2)\ {\itshape Update the dictionary:} Instead of updating the whole dictionary at one time, we renew the dictionary atom by atom to fully utilize those that have already been updated.
In detail, the dictionary is updated in two layers. In the outer layer,
assuming ${\bf D}=[{\bf D}^0,\dots,{\bf D}^C]$ where the part-dictionary ${\bf D}^c \in\mathbb{R}^{M\times K^c}$ is composed of all the atoms with their indices in ${\mathcal{I}^c}$ with the size $K^c$, $c=0,\dots,C$,
part-dictionaries are renewed one by one.
While in the inner layer, for the $c^{th}$ part-dictionary ${\bf D}^c$, the atoms are updated one by one.

Provided $i\in\mathcal{I}^c$ and the other atoms in ${\bf D}$ are fixed, for updating the $i^{th}$ atom ${\bf d}_i$  we arrive at the optimization problem as follows
\begin{align}
\min_{{\bf d}_i}\ \left\| {\bf Y}- \sum_{ k\notin \mathcal{I}^c } {\bf d}_k \bar{\bf x}_k  - \!\!\sum_{k\in \mathcal{I}^c, k\not=i}\!\!\!\!{\bf d}_k\bar{\bf x}_{k} -  {\bf d}_i\bar{\bf x}_{i}\right\|_F^2, \label{eq:updatedi}
\end{align}
where $\bar{\bf x}_k$ denotes the $k^{th}$ row in the whole code matrix ${\bf X}$.
It should be noted that during updating the atoms with indices in $\mathcal{I}^c$, other ones with indices out of $\mathcal{I}^c$  are always unchanged.
Therefore, ${\bf Z} = {\bf Y}- \sum_{ k\notin \mathcal{I}^c } {\bf d}_k \bar{\bf x }_k$ is computed in advance to avoid redundant calculation and the dictionary is updated in two layers.
Problem \eqref{eq:updatedi} is rewritten as
\begin{align}
\min_{{\bf d}_i}\ \left\| {\bf Z}  - \!\!\sum_{k\in \mathcal{I}^c, k\not=i}\!\!\!\!{\bf d}_k{\bf \bar x}_{k} -  {\bf d}_i\bar{\bf x}_i\right\|_F^2. \nonumber
\end{align}
Furthermore, we introduce another variable by
\begin{align}
\tilde{\bf Z}={\bf Z}-\!\!\sum_{k\not= i, k\in \mathcal{I}^c}\!\!\!\!{\bf d}_k \bar{\bf x}_{k},
\label{Y-updateAtom}
\end{align}
and the solution can be easily derived as
\begin{align}
\tilde{\bf d}_i=\frac{1}{\| \bar{\bf x}_i \|_2^2} \tilde{\bf Z}\bar{\bf x}_{i}^{\rm T}.
\label{atomUpdate}
\end{align}
Given the energy of each atom is constrained in (\ref{costv2}), the updated atom is further normalized as
\begin{align}
\hat{\bf d}_i=\frac{\tilde{\bf Z}\bar{\bf x}_{i}^{\rm T}}{\left\| \tilde{\bf Z}\bar{\bf x}_{i}^{\rm T} \right\|_2}.
\label{atomNormalization}
\end{align}
Likewise, we adopt the same procedure as  (\ref{Y-updateAtom}) and  (\ref{atomNormalization}) to update all atoms with indices in $\mathcal{I}^c$.
When the atom update corresponding to $\mathcal {I}^c \ {\rm with}\ c=0,1,\dots,C$ is finished successively in the same way, the whole dictionary is consequently updated.
In addition, when the scalar for the cross-label suppression is very large, the construction for other class samples depends little on the label-particular atoms with indices in $\mathcal{I}^c$ ($c\not=0$). Then we
can only use ${\bf Y}^c$ and ${\bf X}^c$ instead of ${\bf Y}$ and ${\bf X}$ in the above procedures to accelerate the updating for these atoms.
\subsection{Classifier}
After iteratively learning, the learnt dictionary can be used to represent a query image and judge its label.
On the one hand, according to the proposed dictionary structure and the learning model, the large representation coefficients of one query signal on the whole learnt dictionary should be mainly distributed over its closely associated atoms.
The classification scheme namely global coding classifier (GCC) is accordingly proposed.
On the other hand, owing to other label-particular atoms contribute little to its reconstruction,
the signal should also be reconstructed well if using only its corresponding label-particular atoms and the shared ones.
As a result, another classification scheme namely local coding classifier (LCC) is also developed to evaluate our model.

1)\ {\itshape Global coding classifier:} Given a query sample ${\bf y}$ and the learnt dictionary ${\bf D}$, due to its label information is unknown,
its representation without cross-label suppression is obtained
\begin{align}
\hat{\bf x}&= \text{arg} \min_{\bf x} \left\|{\bf y}-{\bf D}{\bf x}\right\|_2^2 + \beta \left\|{\bf x}\right\|_2^2 \nonumber \\
      &= \left({\bf D}^\text{T}{\bf D}+\beta {\bf I}\right)^{-1}{\bf D}^\text{T}{\bf y}.
\end{align}
According to the dictionary structure and the learning algorithm, if the sample $\bf y$ belongs to the $c^{th}$ class,
large coefficients should be mainly concentrated at the atoms with the $c^{th}$ class label as well as the shares ones.
Hence the residual $\left\|{\bf y}-\sum_{k \in {\mathcal{I}^0}\cup{\mathcal{I}^c}} {\bf d}_k\hat{\bf x}(k)\right\|_2^2 $ should be very small and the sum of absolute coefficients
$\sum_{k \in {\mathcal{I}^0}\cup{\mathcal{I}^c}} |\hat{\bf x}(k)|$ should be very large, where $\hat{\bf x}(k)$ represents the $k^{th}$ component of $\hat{\bf x}$.
As a consequence, we can define the following metric for the visual recognition:
\begin{align}
\text{label}({\bf y})=\text{arg}\min_c \frac{\left\|{\bf y}-\!\!\sum\limits_{k \in {\mathcal{I}^0}\cup{\mathcal{I}^c}} \!\!\!{\bf d}_k\hat{\bf x}(k)\right\|_2^2}{\sum\limits_{k \in {\mathcal{I}^0}\cup{\mathcal{I}^c}} \!\!\!|\hat{\bf x}(k)|}.
\end{align}

2)\ {\itshape Local coding classifier:}  We catenate the shared part-dictionary ${\bf D}^0$ with each
label-particular part-dictionary ${\bf D}^c$ together, obtaining $C$ combined part-dictionaries: $\tilde{\bf D}^c = [{\bf D}^0,{\bf D}^i]$  $c=1,2,\cdots, C$.
Further, we force the query sample ${\bf y}$ to be represented by each combined part-dictionary:
\begin{align}
\hat{\bf x}^c&=\text{arg} \min_{\bf x} \left\|{\bf y}-\tilde{\bf D}^c{\bf x}\right\|_2^2+\beta\left\|{\bf x}\right\|_2^2 \nonumber\\
             &= \left({\tilde{\bf D}}^{c^{\rm T}}{\bf D}^c+\beta {\bf I}\right)^{-1}\!\!\!{\tilde{\bf D}}^{c^{\rm T}}{\bf y},\;\qquad c=1,\dots,C.
\end{align}
and the class label can be readily defined as follows

\begin{align}
\text{label}({\bf y}) = \text{arg} \min_{c} \left\|{\bf y}-\tilde{\bf D}^c\hat{\bf x}^c\right\|_2^2.
\end{align}
The better choice can be easily selected from these two classification schemes in the case of supervised learning.
\subsection{Initialization}

We initialize the whole dictionary by the three following steps. First of all, each label-particular part-dictionary ${\bf D}^c$
 is initialized by the relevant training samples with k-means method, $c=1,\dots,C$.
Then, supposing initial label-particular part-dictionary ${\bf D}^c_0 \in \mathbb{R}^{M\times K_c}$, $c=1,\dots,C$,
each class training samples can be reconstructed through their corresponding part-dictionary
\begin{align}
{\bf X}_0^c&={\rm arg}\min_{\bf X} \left\|{\bf Y}^c-{\bf D}_0^c{\bf X}\right\|_F^2+\beta\left\|{\bf X}\right\|_F^2\nonumber\\
           &=\left[ ({\bf D}_0^c)^{\rm T} {\bf D}_0^c +\beta{\bf I}\right]^{-1}({\bf D}_0^c)^{\rm T}{\bf Y}^c \in \mathbb{R}^{H_c\times N_c},\;\qquad c=1,\dots,C,
\end{align}
and the residuals for each class signals are
\begin{align}
{\bf E}^c={\bf Y}^c-{\bf D}_0^c{\bf X}_0^c \in\mathbb{R}^{M\times N^c},\;\qquad c=1,\dots,C.
\label{initResi}
\end{align}
Finally, the shared atoms can be obtained by employing the k-means method with all the residual data ${\bf E}=[{\bf E}^1,\dots,{\bf E}^C]$.
So the dictionary initialization is completed.

As for initializing the representations ${\bf X}$ for the all the training samples ${\bf Y}=[{\bf Y}^1,\cdots,{\bf Y}^C]$, given the initialized dictionary ${\bf D}_0$, they can be obtained as follows
\begin{align}
{\bf X}_0&={\rm arg}\min_{\bf X}\left \|{\bf Y}-{\bf D}_0{\bf X}\right\|_F^2+\beta\left\|{\bf X}\right\|_F^2\nonumber\\
           &=\left( {\bf D}_0^{\rm T} {\bf D}_0 +\beta{\bf I}\right)^{-1}{\bf D}_0^{\rm T}{\bf Y}.
\label{initCode}
\end{align}
Empirically, we find the simple initialization process can work well for visual recognition.

\section{Experiments}
In this section, we will experimentally evaluate our proposed algorithm on a variety of publicly available datasets involving face recognition,
 object category, scenes recognition, texture classification as well as action classification, and compare it with existing works.
\subsection{Experimental Setup}

In the experiments, the proposed approach is compared with developed dictionary methods mainly including SRC \cite{J_Wright_2009}, K-SVD \cite{K-SVD_M_2006},
 Joint\cite{D_Pham_2008}, D-KSVD \cite{DKSVD_2010}, DLSI \cite{DLSI_I_Ramirez_2010}, LC-KSVD \cite{LCKSVD2013} and COPAR \cite{COPAR2014}.
In addition, the well-known classifier linear support machine vector (SVM) \cite{SVM} and other state-of-the-art classification methods for each dataset are also used for comparison.
The parameters $\beta, \gamma$, and $\lambda$ are mainly optimized by 5-fold cross validation on the training set,
and their values as well as the dictionary size for each dataset will be elaborated in the following experimental details.

In order to obtain a stable recognition rate, by default the experiments are implemented over $10$ times training/test splits for each
dataset, unless there are predefined splits or other specific reasons.
Both the averaged recognition accuracy and its standard deviation are reported.

To fairly evaluate the computational efficiency, the same environment is adopted for different algorithms.
Specifically the 64bit Windows 7 operating system and MATLAB2015b are applied in one PC equipped with Intel i7-5930K 3.5GHz CPU, and 32GB memory.
In addition, it should be mentioned that the reported training time includes the cost for initialization and the iterative learning,
and the test time for each query is the average based on testing $100$ queries.

\subsection{Face Recognition}
We validate our proposed algorithms for face recognition on two popular face datasets, Yale face dataset \cite{Yale_face} and the Extended YaleB database \cite{EYaleB}.
Besides we also discuss and analyze the effect of the cross-label suppression and group regularization in this section.
\begin{figure}
  \subfigure[Yale]
    {
      \begin{minipage}[t]{0.5 \textwidth}
      \centering
      \includegraphics[width=7.0cm,height=2.2cm]{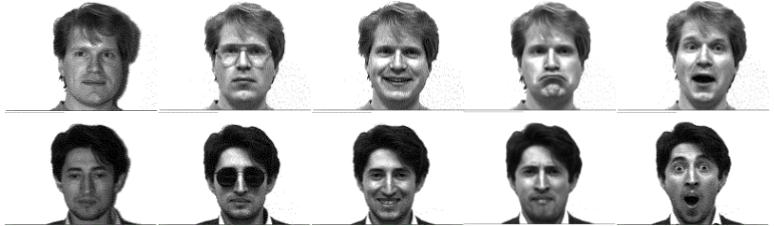}
      \end{minipage}
      \label{Yale-demo}
    }
  \subfigure[Extended YaleB]
      {
      \begin{minipage}[t]{0.5 \textwidth}
      \centering
      \includegraphics[width=7.0cm, height=2.8cm]{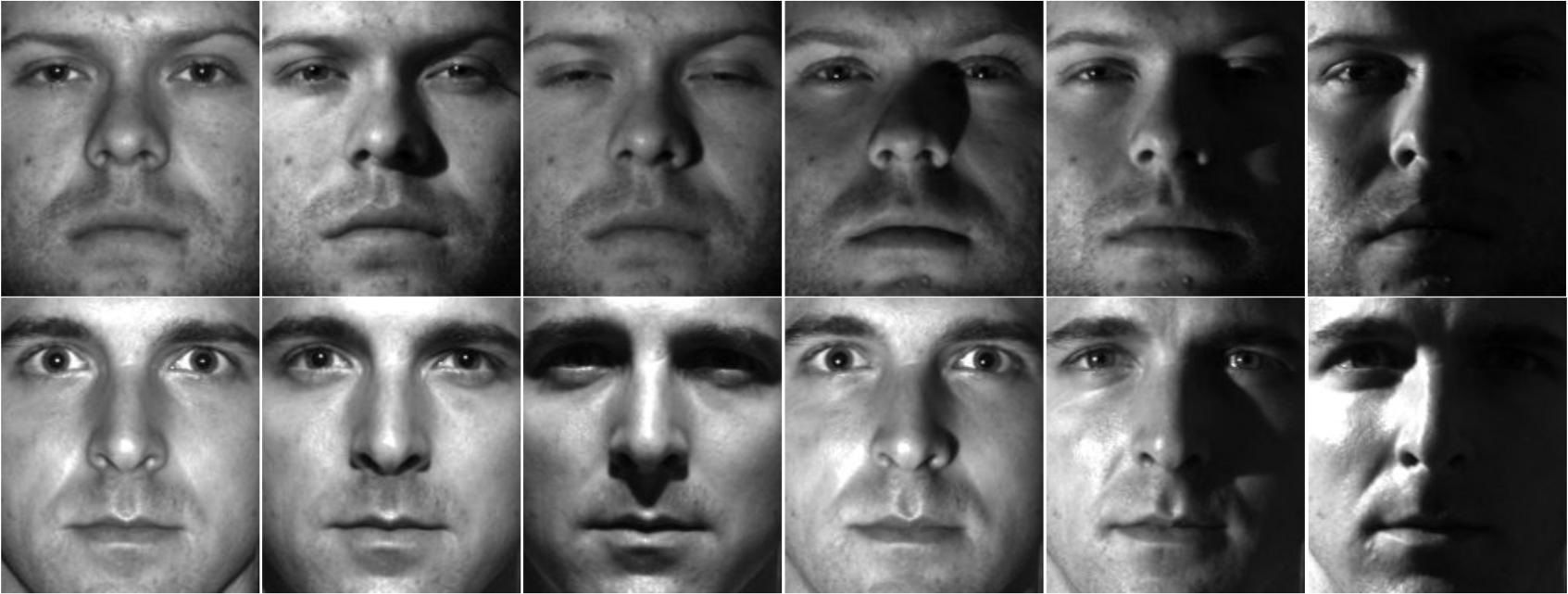}
      \end{minipage}
      \label{EYaleB-demo}
    }
  \vspace{-0.8em}
  \caption{Some example images from the two face datasets: (a) the Yale dataset and (b) the Extended YaleB dataset.}
  \label{face-demo}
\end{figure}

1) {\itshape The Yale face} \cite{Yale_face} : The database contains 165 gray scale images for 15 individuals with 11 images per subject.
For each subject, there's one per different facial expression or configuration: center-light, w/glasses, happy, left-light, w/no glasses, normal, right-light,
sad, sleepy, surprised, and wink, as shown in Figure \ref{Yale-demo}.
All the images are resized to 576-dimensional vectors ($24\times24$-pixel resolution) with normalization for
representation.
Then we randomly select six images of each person for training and keep the rest for testing, using the same experimental
setup as that for COPAR method in \cite{COPAR2014}.
In detail, the dictionary size is set as $15\times4+5=65$ visual atoms, which means 4 label-particular atoms for each person and 5 atoms as the shared.
The parameters $\beta,\ \lambda$, and $\gamma$ are set to $4\times10^{-3}$, $2\times 10^3$, and $1$ in this experiment, respectively .

\begin{table}[t]
\caption{Accuracy on the Yale face dataset}
\label{Yale_comparison}
\centering
\centering
\begin{tabular}{|c|c|c|c|}
\hline
Method&Accuracy\,(\%)&Method&Accuracy\,(\%)\\[0.2ex]
\hline\hline
SVM \cite{SVM}                             &94.42$\,\pm\,2.80$  &DLSI \cite{DLSI_I_Ramirez_2010}        &72.70 \\
SRC \cite{J_Wright_2009}                   &74.60               &LC-KSVD \cite{LCKSVD2013}             &73.60 \\
He \cite{laplacianface_2005}               &88.70               &Wang \cite{F_Wang_2007}                &89.26 \\
Joint \cite{D_Pham_2008}                   &84.61$\,\pm\,4.05$  &COPAR \cite{COPAR2014}                 &78.30 \\
D-KSVD \cite{DKSVD_2010}                   &73.20               &FDDL \cite{FDDL-2014}                  &77.20 \\
\hline
ours\,(LCC)                  &93.65$\,\pm \,3.26$ &{\bf ours\,(GCC)}              &{\bf 95.92}$\,{\bf \pm}\, {\bf 2.23}$\\
\hline
\end{tabular}
\end{table}

Due to there are lots of variations in terms of facial expression and this dataset is also small, the results fluctuate obviously with relatively large standard deviations.
Hence, the experiment results for our algorithm listed in Table \ref{Yale_comparison} are average over $50$ rather than $10$ times independent training/testing splits.
The reported results for this dataset in \cite{COPAR2014} are also included for comparison.
As shown in the table, the global coding classifier outperforms all the comparing methods including the COPAR \cite{COPAR2014} method with the same dictionary structure as ours,
and the local coding classifier also get a competitive result.
Moreover, the best result attained by the GCC, is $1.5\%$ higher than the linear SVM and about $6.5\%$ higher than recognition rate reported in \cite{F_Wang_2007},
which adopted high-level features, spectral ones rather than the plain pixel information.

2) {\itshape The Extended YaleB face }\cite{EYaleB} : That database contains $2,414$ frontal face images of $38$ people. There are about $64$ images for each person.
The original images were cropped to $192 \times 168$ pixels.
This database is challenging due to varying illumination conditions and poses,
as shown in Figure \ref{EYaleB-demo}.
In this database, the Eigenface feature \cite{Eigenfaces} with dimension $300$ from normalized $54\times 48$ images is used instead of the original pixel information for its better performance.
Besides randomly selecting half of the images per category as training, we further evaluate our method under more difficult
conditions with fewer training samples for learning, that is, 20 or even 10 training samples per subject, and the corresponding remaining samples are
used as the test data.
For this dataset, we only apply label-particular atoms and don't adopt any shared atoms.
In the case of $10$ training samples for each person,
the dictionary size is set as $10\times38=380$, which indicates $10$ label-particular atoms for each subject.
Moreover, for the rest two cases, the dictionary sizes are set as $15\times38=570$ with $15$ label-particular atoms for each subject.
The dictionary sizes for other compared dictionary methods are kept the same as ours,
including K-SVD \cite{K-SVD_M_2006}, Joint \cite{D_Pham_2008}, D-KSVD \cite{DKSVD_2010}, DLSI \cite{DLSI_I_Ramirez_2010}, LC-KSVD \cite{LCKSVD2013}, COPAR \cite{COPAR2014}.
The parameters $\beta, \lambda$, and $\gamma$ are optimized as $8\times10^{-3}$, $2\times 10^3$, and $2\times10^{-2}$ in this experiment, respectively.
In addition, non-dictionary classification methods like the sophisticated classifier linear support vector machine (SVM),
and one general classifier $k$-nearest neighbors classifier (k-NN) with the Euclidean distance are also used.
The average results and standard deviations are shown in Table \ref{EYale_comparison}, in which related reported results in \cite{CRC_2011}, \cite{RCR_2012}, \cite{FDDL_D_Yang2011} are also listed.

\begin{table}
\caption{Accuracy on Extended YaleB face dataset with Eigenface}
\label{EYale_comparison}
\centering
\begin{tabular}{|c|c|c|c|}
\hline
{num of train samp.}&10&20&32\\
\hline
\hline
Method&\multicolumn{3}{c|}{Accuracy\,(\%)}\\[0.2ex]
\hline
SVM \cite{SVM}                   &84.56$\,\pm\,1.57$  &92.54$\,\pm\,0.82$          &96.42$\,\pm\,0.47$\\
k-NN                             &57.88$\,\pm\,1.40$  &73.72$\,\pm\,1.37$          &80.86$\,\pm\,0.85$\\
SRC \cite{J_Wright_2009}         &89.66$\,\pm\,0.96$  &95.38$\,\pm\,0.72$          &97.64$\,\pm\,0.43$\\
K-SVD \cite{K-SVD_M_2006}        &77.12$\,\pm\,1.88$  &89.55$\,\pm\,1.15$          &92.95$\,\pm\,1.31$\\
Joint \cite{D_Pham_2008}         &86.56$\,\pm\,1.17$  &91.21$\,\pm\,0.88$          &94.28$\,\pm\,0.78$\\
D-KSVD \cite{DKSVD_2010}          &80.53$\,\pm\,1.21$  &88.77$\,\pm\,0.97$          &94.39$\,\pm\,0.67$\\
DLSI \cite{DLSI_I_Ramirez_2010}  &83.08$\,\pm\,1.22$  &91.36$\,\pm\,0.97$          &94.04$\,\pm\,0.84$\\
LC-KSVD \cite{LCKSVD2013}        &81.57$\,\pm\,1.84$  &92.36$\,\pm\,1.41$          &94.92$\,\pm\,0.61$\\
COPAR \cite{COPAR2014}           &89.55$\,\pm\,1.01$  &95.82$\,\pm\,0.88$          &97.33$\,\pm\,0.43$\\
FDDL \cite{FDDL-2014}            &-                   &92.40                       &-         \\
CRC \cite{CRC_2011}              &84.80               &91.20                       &97.90         \\
RCR \cite{RCR_2012}              &86.80               &92.30                       &-             \\
LRC \cite{LRC_2010}              &82.40               &87.00                       &95.90         \\
{\bf ours\,(GCC)}                &{\bf 93.38}$\,\pm\, {\bf0.76}$     &{\bf 97.46}$\,\pm\,{\bf 0.56}$  &{\bf 98.62}$\,\pm\,{\bf 0.38}$\\
\hline
\end{tabular}
\end{table}

We can see our method always get the best accuracy compared with competing methods in all the cases.
Additionally even for fewer training samples ($20$ per subject), satisfying recognition rate of $97.46\%$ is attained,
which is even much higher than the accuracies obtained by the linear SVM \cite{SVM} and some developed dictionary methods with $32$ training samples for each subject, such as Joint \cite{D_Pham_2008}, D-KSVD \cite{DKSVD_2010}, DLSI \cite{DLSI_I_Ramirez_2010}, and LC-KSVD \cite{LCKSVD2013}.
In terms of standard deviations, our approach perform more stably than k-NN, the linear SVM \cite{SVM}, and other dictionary methods.
From our proposed algorithm, it can be found that fewer training samples are, and larger the attained standard deviation is.
This observation is also shared by other methods such as SRC \cite{J_Wright_2009}, D-KSVD \cite{DKSVD_2010}, COPAR \cite{COPAR2014}, LC-KSVD \cite{LCKSVD2013}, and so on.
It demonstrates that when the training set is smaller, which samples the training set consists of plays a more important role for classification.

Additionally, to directly compare with the graph regularized dictionary learning approach \cite{supGraphDL-arxiv2016},
we also investigate random-faces with 504 dimension for our algorithm in the case of $32$ training samples for each subject, following its experiment.
All the parameters are simply set to the same values as those for the Eigenface features without any fine-tuning.
The recognition accuracy for our approach is listed in Table \ref{EYale_randomface},
and results reported in the recent work \cite{supGraphDL-arxiv2016} for LC-KSVD \cite{LCKSVD2013}, SupGraphDL-L \cite{supGraphDL-arxiv2016} with random training/test splits are also included for comparison.
Apparently, our algorithm performs the best with a significant improvement.
Compared with LC-KSVD \cite{LCKSVD2013}, SupGraphDL-L \cite{supGraphDL-arxiv2016} adopts adaptive graph regularization for dictionary learning and sparse coding individually, and then the enhancement is attained.

\begin{table}
\caption{Accuracy on Extended YaleB face dataset with random-face}
\label{EYale_randomface}
\centering
\begin{tabular}{|c|c|c|c|}
\hline
{Method}&{Accuracy\,(\%)}&{Method}&{Accuracy\,(\%)}\\
\hline\hline
SupGraphDL\,\cite{supGraphDL-arxiv2016}    &92.89    &LC-KSVD\,\cite{LCKSVD2013}                 &93.29  \\
SupGraphDL-L\,\cite{supGraphDL-arxiv2016}  &93.44    &{\bf GCC(ours)}                           &{\bf 98.49}$\,\pm\,{\bf 0.39}$\\
\hline
\end{tabular}
\end{table}

The computational efficiency comparison is listed in Table \ref{EYaleB-computation},
and it can seen that our method performs the fastest for training dictionary with almost $5$ times and $28$ times as fast as LC-KSVD \cite{LCKSVD2013} and COPAR \cite{COPAR2014}, respectively.
For classifying one query, our approach is also the fastest, and outperforms COPAR \cite{COPAR2014},
and SRC \cite{J_Wright_2009} by two orders of magnitude.
Due to in LC-KSVD \cite{LCKSVD2013} one linear classifier is jointly trained with the dictionary, it's very fast for testing.
Nonetheless, our approach still possesses a marginal advantage over it for its fast coding.

\begin{table}[t]
\caption{computation efficiency on Extended YaleB face database with $32$ training samples for each subject}
\label{EYaleB-computation}
\centering
\begin{tabular}{|c|c|c|}
\hline
Method&Training time\,(s)&Time per test samp.\,(ms)\\
\hline\hline
SRC \cite{J_Wright_2009}      &-      &31.10\\
LC-KSVD \cite{LCKSVD2013}     &57.91  &0.32\\
COPAR \cite{COPAR2014}        &338.36 &22.21\\
\hline
{\bf ours\,(GCC)}               &{\bf 12.66}  &{\bf 0.24}\\
\hline
\end{tabular}
\end{table}

\begin{figure}[!t]
\centering
  \subfigure[]
  {
      \begin{minipage}[]{0.6 \textwidth}
      \label{EYaleB-lambda}
      \centering
      \includegraphics[width=10cm]{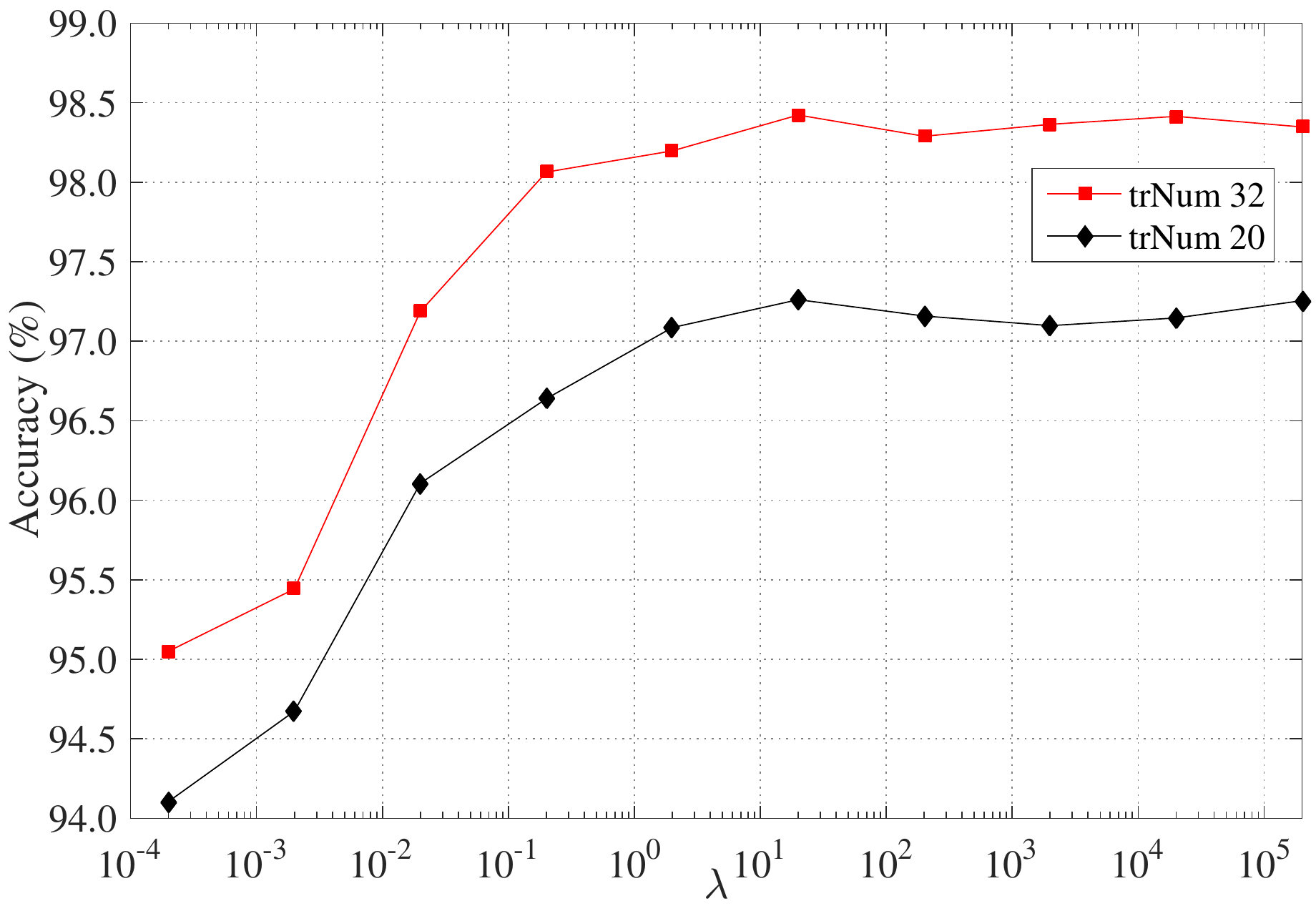}
      \end{minipage}
  }
\\
  \noindent\subfigure[]
  {
      \begin{minipage}[]{0.6 \textwidth}
      \label{EYaleB-gammma}
      \centering
      \includegraphics[width=10cm]{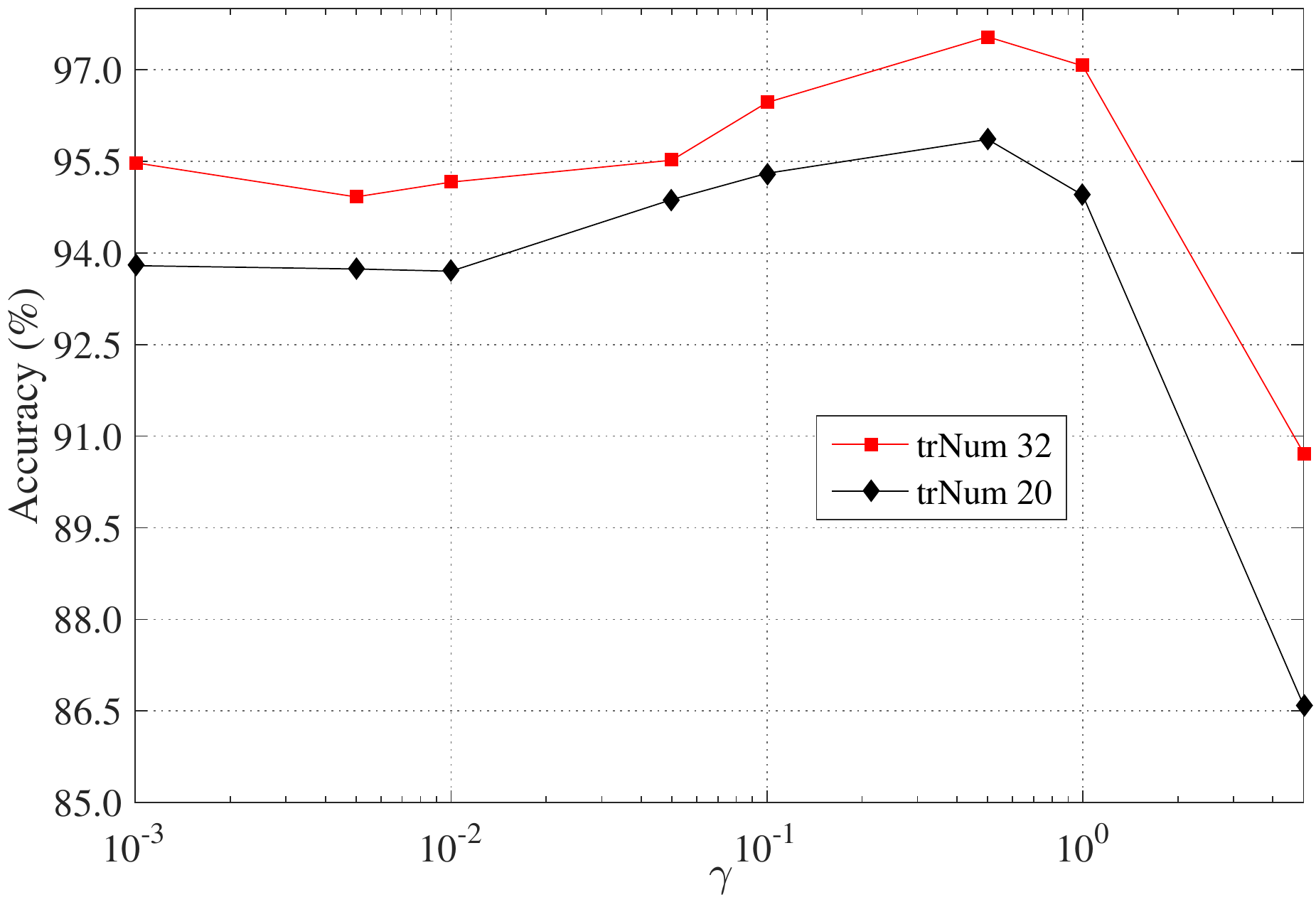}
      \end{minipage}
    }
  \vspace{-1em}
  \caption
  {
  The effect of the cross-label suppression and group regularization on performance for the Extended YaleB .
  (a) The effect of the cross-label suppression with $\gamma=0$ in the cases of {\itshape trNum} $= 20$ and {\itshape trNum} $=32$.
  (b) The effect of the group regularization  with $\lambda=0$ in the cases of {\itshape trNum} $= 20$ and {\itshape trNum} $=32$.
  }
\end{figure}

3)  {\itshape The effect of the cross-label suppression and group regularization}:
To understand the effect of the cross-label suppression and group regularization, we conduct two experiments on the Extended YaleB dataset by varying one scalar while fixing the other scalar at zero, with other experimental conditions kept the same.
The results with various scalars $\lambda$ without the group regularization for is shown in
Figure \ref{EYaleB-lambda}.
It can founded that the curves for two cases of training samples look very similar.
Moreover when $\lambda$ is more than $20$, very significant improvements are obtained and then plateaus almost appear.
According to the effect curves, the set $\{2\times10^{-1},\ 2,\ 2\times10^1,\ 2\times 10^2,\ 2\times 10^3\}$ can be considered as a rough range for $\lambda$ selection in other datasets.
Further, the effect on classification accuracy of the group regularization without the cross-label suppression is shown in Figure \ref{EYaleB-gammma}.
The obvious improvements brought by the group regularization are about $2.0\%$ and $2.1\%$ severally in the cases of {\itshape trNum} $= 20$ and {\itshape trNum} $=32$ with {\itshape trNum}
denoting the number of training samples for each class.
Therefore, the significance of those two terms is well demonstrated.

\begin{figure}[!t]
\centering
  \subfigure[subject 16]{
      \begin{minipage}[t]{0.7 \textwidth}
      \flushleft
      \includegraphics[width=10cm]{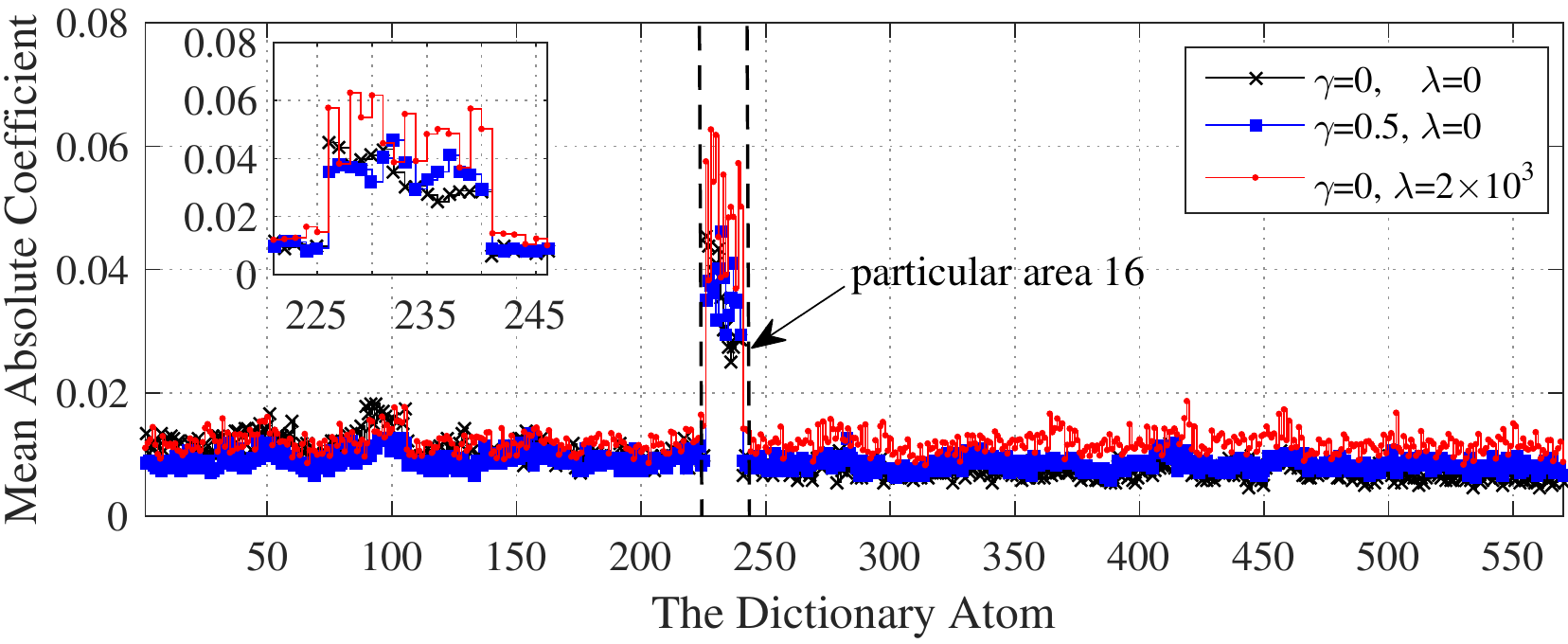}
      \end{minipage}
    }
    \\
  \subfigure[subject 26]{
      \begin{minipage}[t]{0.7 \textwidth}
      \flushleft
      \includegraphics[width=10cm]{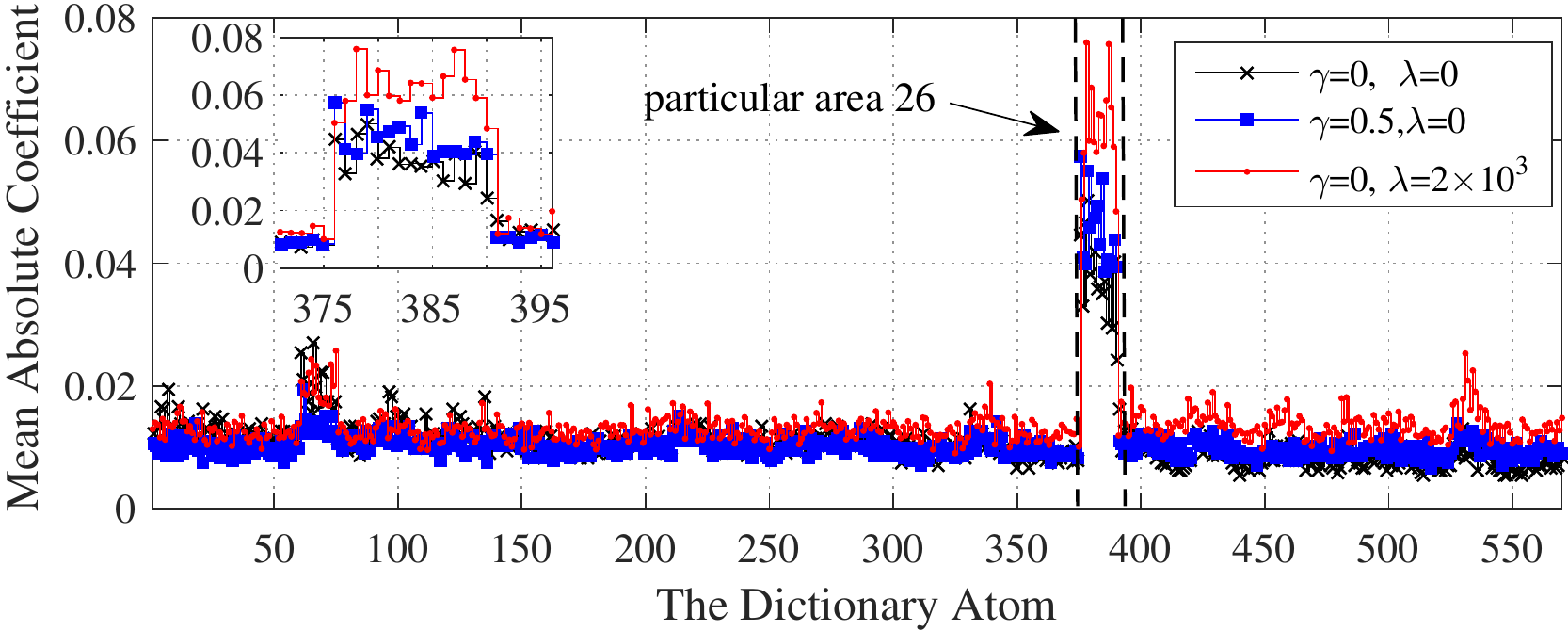}
      \end{minipage}
    }
    \\
    \subfigure[subject 36]{
      \begin{minipage}[t]{0.7 \textwidth}
      \flushleft
      \includegraphics[width=10cm]{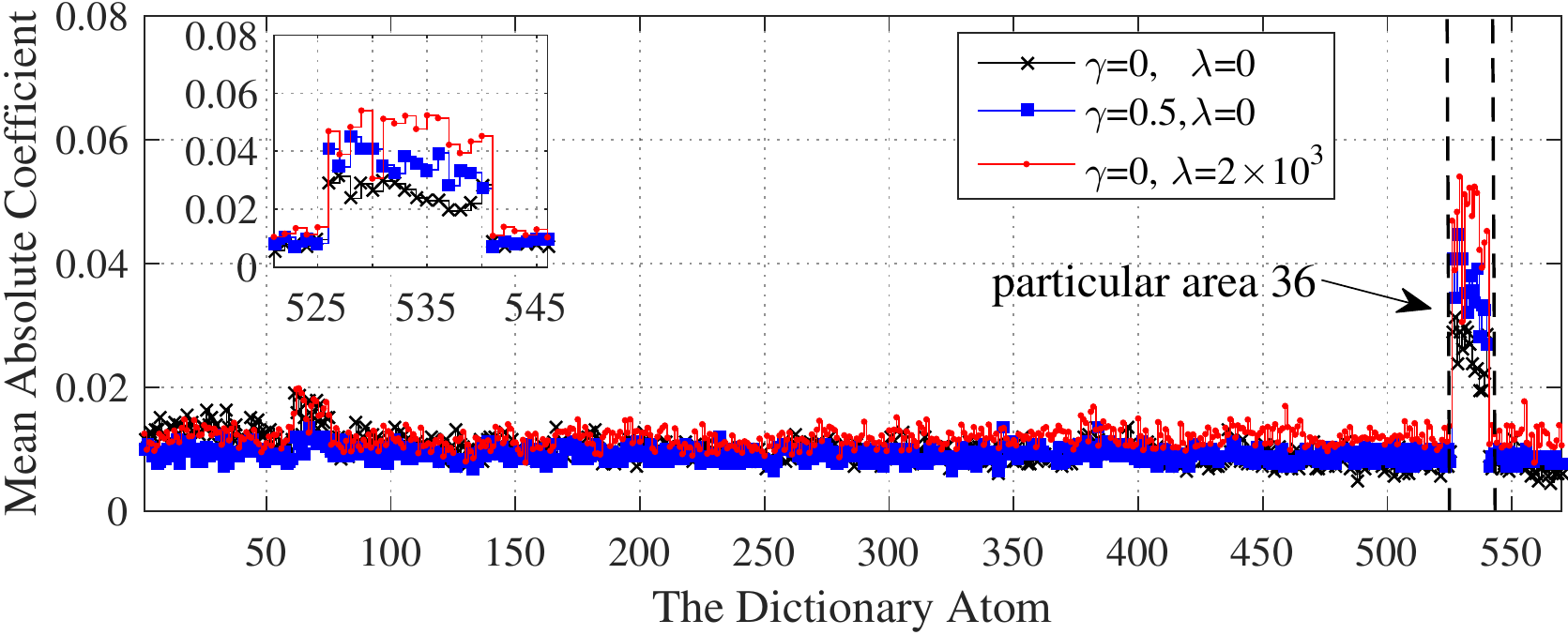}
      \end{minipage}
    }
  \vspace{-0.5em}
  \caption{Each waveform indicates the mean absolute representation over the dictionary for all the testing samples from the same class, and the red, blue and black curves in each subfigure
  corresponding to some class denote the results in the three cases of $\gamma$ and $\lambda$.  Besides the particular area comprise all the atoms with the same particular label.}
  \label{EYaleB-blockStructure}
\end{figure}

To investigate the reason for the enhancement brought by the cross-label suppression and the group regularization,
for the Extended YaleB database with $32$ training samples per class,
we compare the representation distributions over the learnt dictionaries in the three cases with $\{\gamma=0,\ \lambda=0\}$
, $\{\gamma=5\times 10^{-1},\ \lambda=0\}$, and $\{\gamma=0,\ \lambda=2\times10^3\}$ individually,
while other conditions are kept the same, shown in Figure \ref{EYaleB-blockStructure}.
The case with $\{\gamma=0,\ \lambda=0\}$ is viewed as the baseline.
It can be found that
for each class, the representation coefficients locating in its own particular area become larger for both the cases of group regularization and cross-label suppression,
and then the block structures become more distinguished.
Therefore, the representations for different classes become more discriminative, and the better performances for classification are consequently obtained for the two regularization cases.
Besides, the most distinct block structures are brought by the cross-label suppression, and they account for the best results attained by it.

Interestingly, it's should be noted that without the cross-label suppression, block structures are also exhibited over the learnt dictionary.
The phenomenon can be attributed to
the distinct initialization method that each part-dictionary is respectively initialized using the training samples of the corresponding class in the whole dictionary.
When the dictionary is wholly initialized with randomly selected samples from the training set,
as shown in Figure \ref{wholeInit}, block structures disappear for the cases without the cross-label suppression.
However, even with a small scalar for the cross-label suppression, the block structure is still obviously obtained.
Compared with the distinct initialization, the recognition accurate is slightly decreased from 98.06 to 97.46 when $\gamma=0$ and $\lambda=0.2$.

\begin{figure}[!t]
\begin{center}
\includegraphics[width=10cm]{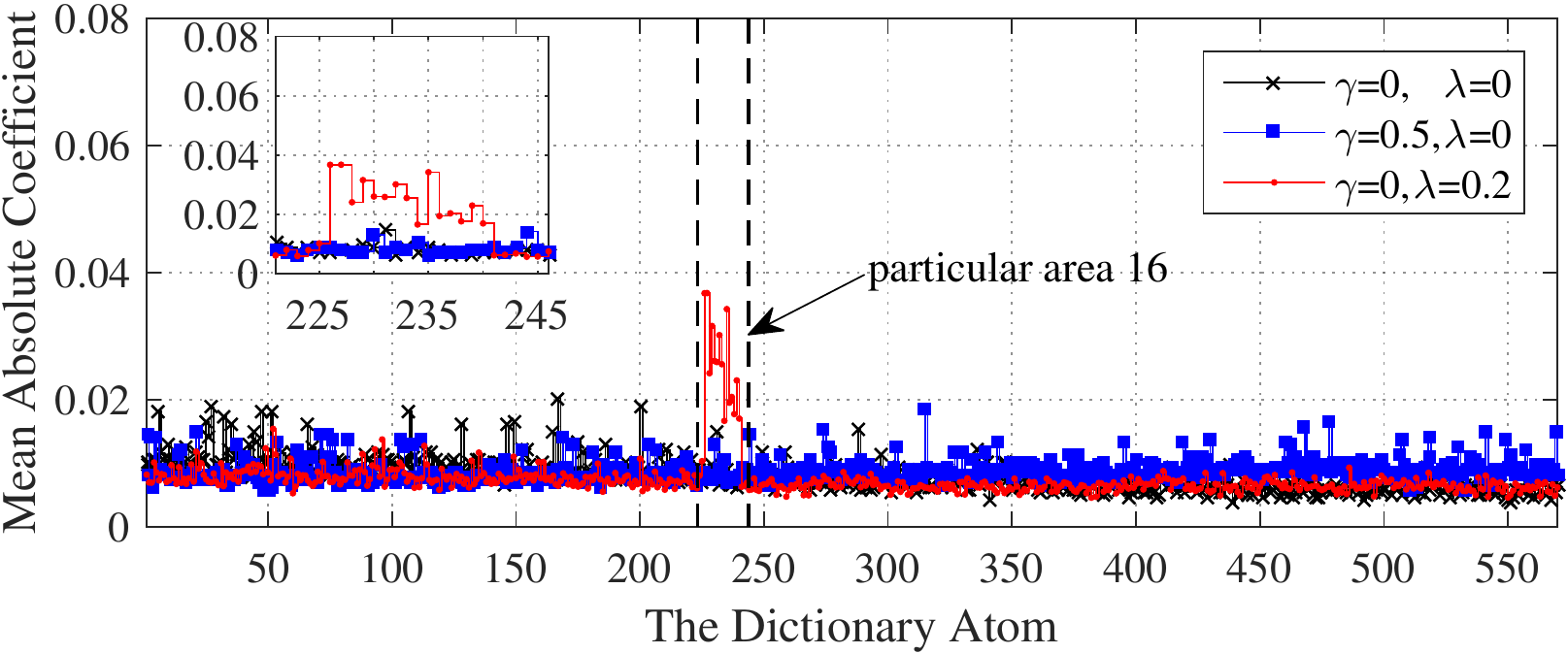}
\caption{Each waveform indicates the mean absolute representation over the dictionary for all the testing samples from the class 16, with the dictionary wholly initialized by randomly selected samples.}
\label{wholeInit}
\end{center}
\end{figure}

\subsection{Caltech101}

\begin{table}[t]
\caption{Accuracy on Caltech101 dataset}
\label{Caltech101-comparison}
\centering
\begin{tabular}{|c|c|c|c|c|}
\hline
{num of train samp.}&15&20&25&30\\
\hline
\hline
Method&\multicolumn{4}{c|}{Accuracy(\%)}\\[0.2ex]
\hline
Malik \cite{Malik-2006}                      & 59.10    &62.00     &-      &66.20\\
Lazebnik \cite{S_Lazebink_2006}              & 56.40    &-         &-      &64.60\\
Griffin \cite{Griffin-2007}                  & 59.00    &63.30     &65.80  &67.60\\
Irani \cite{Irani-2008}                      & 65.00    &-         &-      &70.40\\
Grauman \cite{Grauman-2008}                  & 61.00    &-         &-      &69.10\\
Gemert \cite{Gemert_2008}                    & -        &-         &-      &64.16\\
Yang \cite{Yang-LSPM2009}                    &67.00     &-         &-      &73.20\\
Spanias \cite{Spanias-2011}                  &64.73     &67.96     &70.28  &72.40\\
LLC \cite{J_Wang_LLC2010}                    &65.43     &67.74     &70.16  &73.44\\
SVM \cite{SVM}                               &66.31     &68.53     &70.71  &71.98\\
SRC \cite{J_Wright_2009}                     &64.90     &67.70     &69.20  &70.70\\
K-SVD \cite{K-SVD_M_2006}                    &65.20     &68.70     &71.00  &73.20\\
Joint \cite{D_Pham_2008}                     & 42.00    &-         &-      &-\\
D-KSVD \cite{DKSVD_2010}                     &65.10     &68.60     &71.70  &73.00\\
DLSI \cite{DLSI_I_Ramirez_2010}              &60.90     &65.34     &67.89  &70.34\\
LC-KSVD \cite{LCKSVD2013}                    &67.70     &70.50     &72.30  &73.60\\
COPAR \cite{COPAR2014}                       &62.31     &66.78     &69.82  &71.75  \\
{\bf ours\,(LCC)}                            &{\bf 70.51}     &{\bf 73.65}     &{\bf 76.18}  &{\bf 77.94}  \\
\hline
\end{tabular}
\end{table}

Caltech101 dataset \cite{caltech101-dataset} is a challenging dataset for object recognition with a large number of classes (i.e. $101$ object classes and one background class), including animals, faces, vehicles, flowers, insects and so on.
The dataset totally contains $9,144$ images with roughly $300\times200$ pixels.
For a reasonable comparison, we also adopt the 3000-dimensional SIFT-based features used in LC-KSVD \cite{LCKSVD2013} for which a four-level spatial pyramid and PCA are applied, in our experiments.
Following the common experimental settings, $15$, $20$, $25$, $30$
samples per category are randomly selected for training with testing on the rest samples.
Given {\itshape trNum} training samples for each category, we apply {\itshape trNum} - 1 label-particular atoms for each category and $100$ shared atoms for all the categories in our dictionary,
with the same structure always adopted by COPAR \cite{COPAR2014}.
The parameters $\beta,\ \lambda$, and $\gamma$ are set to $2, 2\times 10^2$, and $10$ in the experiments, respectively.
Likewise our experiments are repeated $10$ times using different training/test splits for reliable accuracies.
Our results are compared with dictionary based methods K-SVD \cite{K-SVD_M_2006}, Joint \cite{D_Pham_2008}, D-KSVD \cite{DKSVD_2010}, DLSI \cite{DLSI_I_Ramirez_2010}, LLC \cite{J_Wang_LLC2010}, LC-KSVD \cite{LCKSVD2013}, and \cite{COPAR2014} and other state-of-the-art approaches, and listed in Table \ref{Caltech101-comparison}.

We can find that our method outperforms the other competing approaches in all the $4$ training cases, and always outdoes the second best method LC-KSVD \cite{LCKSVD2013} by around $3\%$.
Although the commonality is also adopted for COPAR \cite{COPAR2014} like our method, compared with it, recognition advantage of our approach is very distinguished.
That can be well explained by the gain brought by the cross-label suppression and group regularization, illustrated by Figure \ref{Caltech101-gamma-lambda}.
It also can be seen that when $\lambda$ is fixed at $2\times 10^2$, for different training cases,
the performance curves with respect to various scalars of the group regularization resemble one another, and they indicate the relatively optimal $\gamma$ can be shared by diverse training case for the same dataset.

\begin{figure}[t]
\centering
  \subfigure[$\lambda =0$]{
      \begin{minipage}[t]{0.7 \textwidth}
      \flushleft
      \includegraphics[width=10cm]{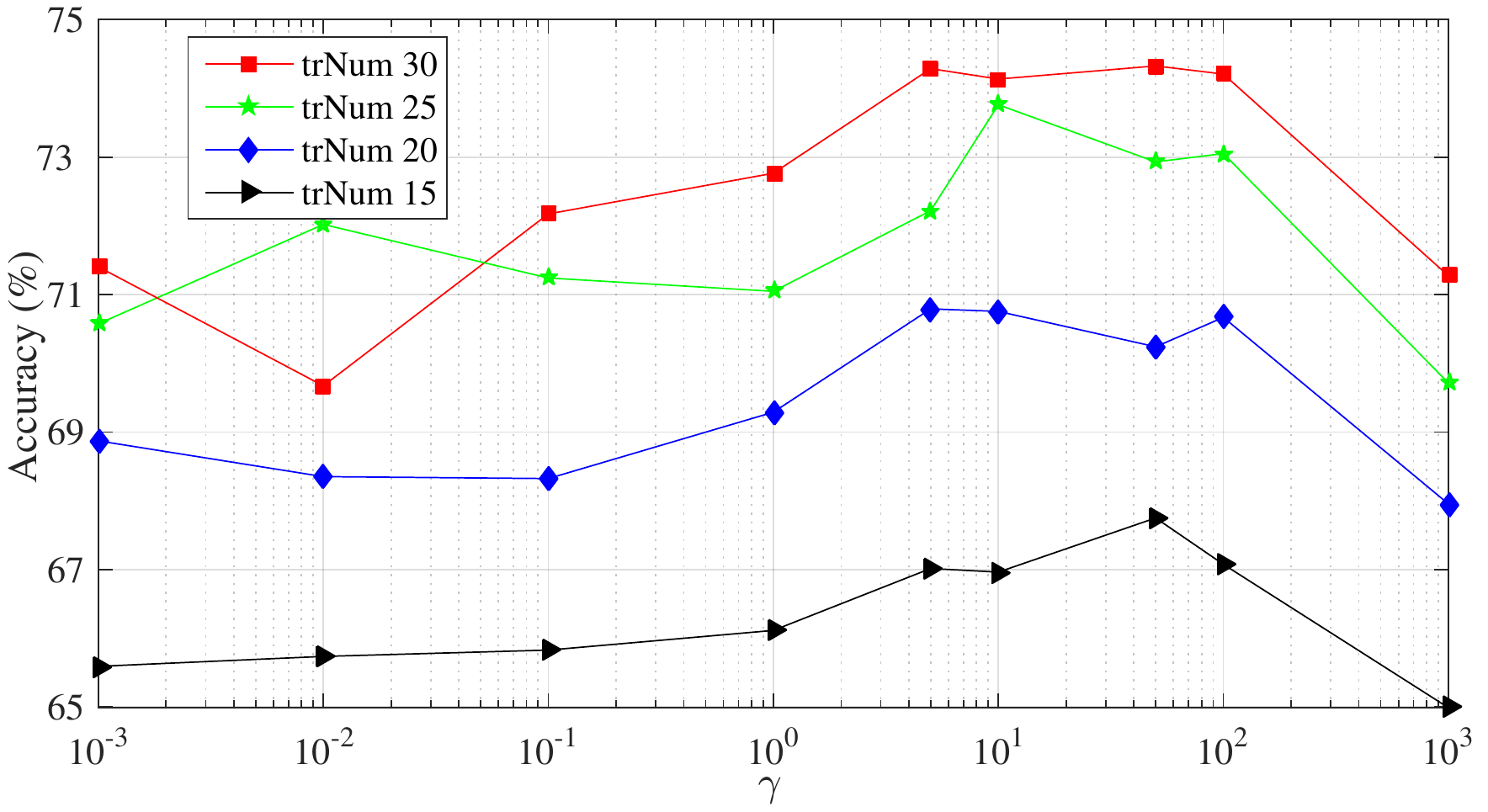}
      \end{minipage}
    }
    \\
  \subfigure[$\lambda=2\times 10^2$]{
      \begin{minipage}[t]{0.7 \textwidth}
      \flushleft
      \includegraphics[width=10cm]{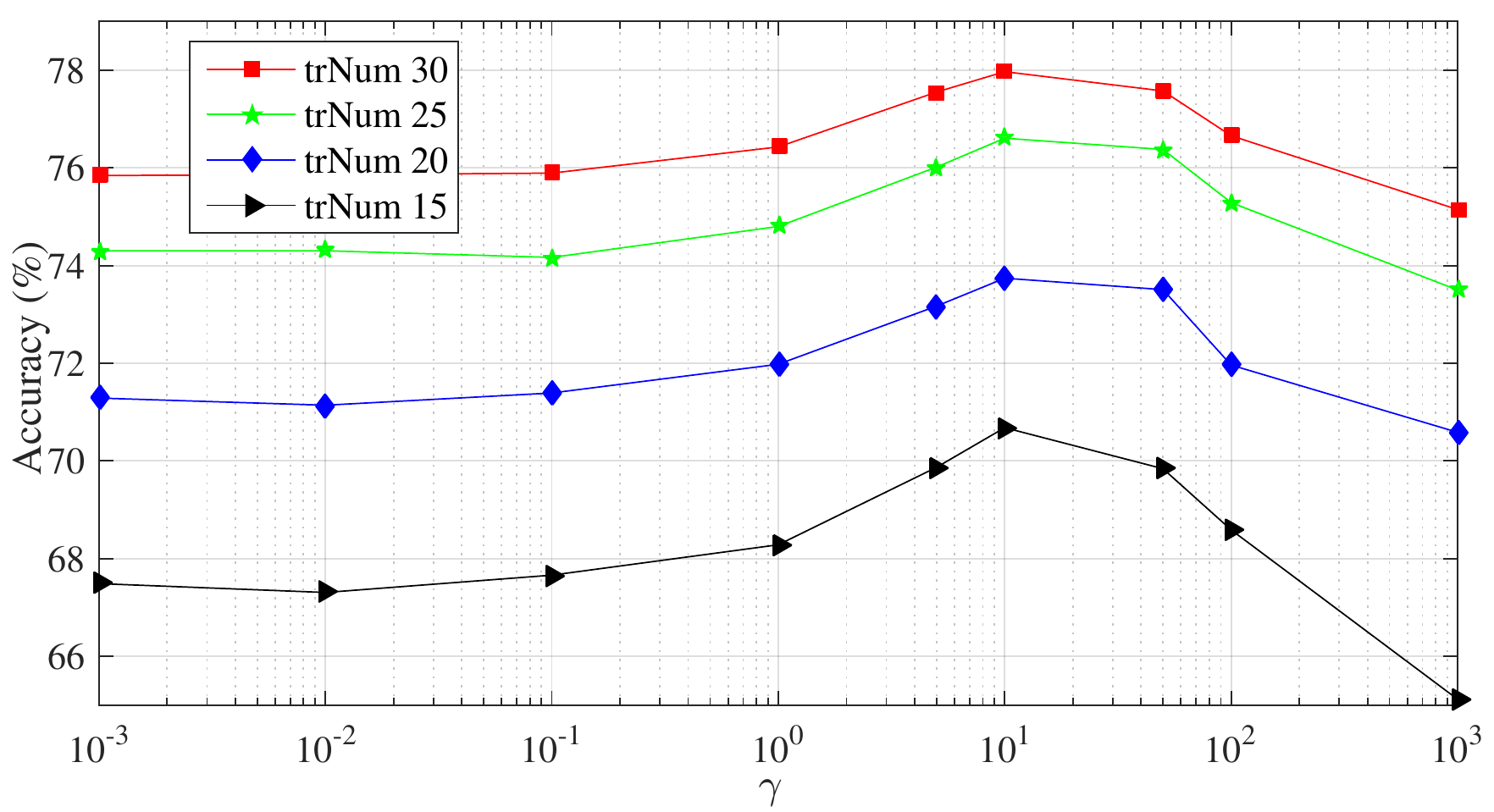}
      \end{minipage}
    }
  \caption{The effect on the Caltech101 classification accuracy of the group regularization with $\lambda=0$ and $\lambda=2\times 10^2$.}
  \label{Caltech101-gamma-lambda}
\end{figure}

\begin{table}[t]
\centering
\begin{subtable}
\centering
\caption{ computation time for training on Caltech101}
\label{Caltech101-training}
\begin{tabular}{|c|c|c|c|c|}
\hline
{num of train samp.}&15&20&25&30\\
\hline
\hline
Method&\multicolumn{4}{c|}{Training time\,(s)}\\[0.2ex]
\hline
{LC-KSVD} \cite{LCKSVD2013}               & {314.00}      &{1006.71}     &{ 1903.31}   &{ 3238.82}\\
COPAR \cite{COPAR2014}                    & 2195.42       & {4208.29}    &{7132.87}    &{11262.42}\\
\hline
{\bf ours\,(LCC)}                         &{\bf 45.41}    &{\bf 82.90}   &{\bf 136.72}  &{\bf 213.85}\\
\hline
\end{tabular}
\end{subtable}

\begin{subtable}
\centering
\caption{Computation time for testing one sample on Caltech101}
\label{Caltech101-test}
\begin{tabular}{|c|c|c|c|c|}
\hline
{num of train samp.}&15&20&25&30\\
\hline
\hline
Method&\multicolumn{4}{c|}{Time per test samp.\,(ms)}\\[0.2ex]
\hline
SRC \cite{J_Wright_2009}                      & 104.90    &172.12   &227.10 &315.20\\
{\bf LC-KSVD} \cite{LCKSVD2013}               & {\bf 1.51}      &{\bf 2.30}     &{\bf 2.80}   &{\bf 4.80}\\
COPAR \cite{COPAR2014}                        & 117.70    &167.76   &240.56 &311.80\\
\hline
ours\,(LCC)                                   &8.40       &8.70     &9.02     &9.41\\
\hline
\end{tabular}
\end{subtable}
\end{table}

The computation efficiency comparison for training is listed in Table \ref{Caltech101-training},
and it can founded that our method performs the most efficiently for learning dictionary in all the training cases,
with about $10$ times and $50$ times faster than LC-KSVD \cite{LCKSVD2013} and COPAR \cite{COPAR2014}.
From computation efficiency comparison for testing listed in Table \ref{Caltech101-test}, we can see that due to we adopt local coding classifier and it needs coding over each combined part-dictionary,
our approach performs slightly slower for categorization than LC-KSVD \cite{LCKSVD2013}, which benefits from its learnt linear classifier.
However, the testing time of our approach is still very small and it's $12\sim33$ times faster than SRC \cite{J_Wright_2009} and COPAR \cite{COPAR2014}.

\subsection{Scene 15}
This dataset of 15 natural scene categories introduced in \cite{scene15_2007}, contains a wide range of outdoor and indoor scene environments such as office, bedroom,
industrial, tall building, mountain and suburb, shown in Figure \ref{Scene15-samp}.
Each category has 200 to 400 images with the total number of $4485$, and the average image size is about $250 \times 300$ pixels.
For a fair comparison, like the case in Caltech101 dataset, we also use the 3000-dimensional SIFT-based features applied by LC-KSVD \cite{LCKSVD2013}.
Following the common experimental settings, we randomly select 100 images per category as training data and use the rest for testing.
The learned dictionary has $15\times30=450$ atoms without any shared atoms, and $10$ times independent training/test splits are implemented.
The parameters $\beta$ and $\lambda$ are optimized as $2\times 10^{-3}$ and $2\times 10^{-1}$, and $\gamma$ for LCC and GCC is respectively optimized as $10$ and $1$.

Beside dictionary-based methods like K-SVD \cite{K-SVD_M_2006}, Joint \cite{D_Pham_2008}, D-KSVD \cite{DKSVD_2010}, DLSI \cite{DLSI_I_Ramirez_2010}, LLC \cite{J_Wang_LLC2010},
LC-KSVD \cite{LCKSVD2013}, and COPAR \cite{COPAR2014} with the same dictionary structure as ours,  we also compare our approach with other state-of-the-art scene classification methods
\cite{scene15_2007,Gemert_2008,Yang-LSPM2009,Gao_2010,Lian-2010,Boureau-2010}.
From the Table \ref{Scene15_comparison}, it can be seen that both LCC and GCC of our approach outperform competing methods by a significant improvement,
and LCC has a marginal advantage over GCC by $0.36\%$ for this dataset.
The confusion matrices with LCC and GCC are shown in Figure \ref{Scene15-fig}
in which dominant diagonals are well-marked. It should be noted that confusion among classes is very little and both LCC and GCC attain more than $99.0\%$
recognition rate for suburb, highway, inside-city, and street, office.
It also can be seen that the slight superiority of LCC to GCC is mainly attributed to its better performance in the industrial, kitchen and store categories.

\begin{figure}
\begin{center}
\includegraphics[width=10cm]{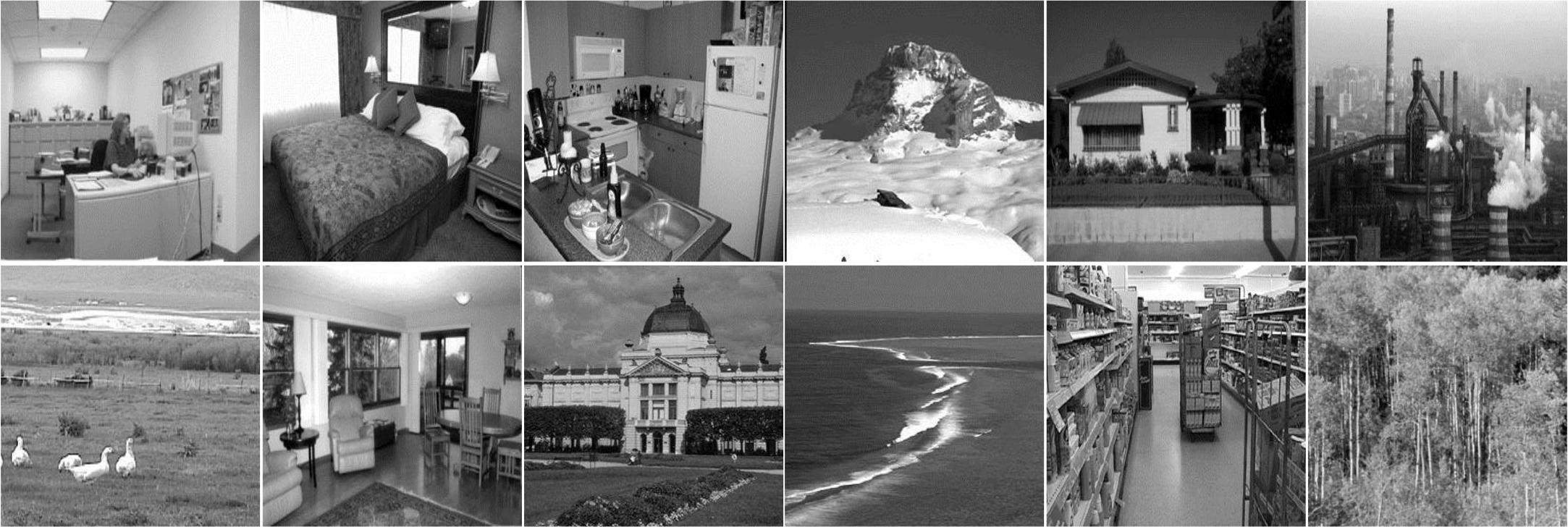}
\caption{Examples from 12 categories in Scene15 dataset.}
\label{Scene15-samp}
\end{center}
\end{figure}

\begin{table}[t]
\caption{Accuracy on Scene15 dataset}
\label{Scene15_comparison}
\centering
\begin{tabular}{|c|c|c|c|}
\hline
Method&Accuracy\,(\%)&Method&Accuracy\,(\%)\\[0.2ex]
\hline\hline
Lazebink \cite{scene15_2007}    &81.40            &SVM \cite{SVM}                     &95.06$\,\pm\,{0.47}$\\
Gemert \cite{Gemert_2008}       &76.70            &SRC \cite{J_Wright_2009}           &91.80\\
Yang \cite{Yang-LSPM2009}       &80.30            &K-SVD \cite{K-SVD_M_2006}          &86.70\\
Gao \cite{Gao_2010}             &89.70            &COPAR \cite{COPAR2014}             &95.54$\,\pm\,{0.41}$\\
Lian \cite{Lian-2010}           &86.40            &D-KSVD \cite{DKSVD_2010}           &89.10\\
Boureau \cite{Boureau-2010}     &84.30            &Joint \cite{D_Pham_2008}           &88.20\\
LLC \cite{J_Wang_LLC2010}       &89.20            &DLSI \cite{DLSI_I_Ramirez_2010}    &92.46\\
Liu \cite{ASP-2016}             &84.70            &LC-KSVD \cite{LCKSVD2013}          &92.90\\
\hline
{\bf ours\,(GCC)}             &{\bf 98.30}$\,\pm\, {\bf 0.26}$   &{\bf ours\,(LCC)}                  &{\bf 98.66}$\,\pm\,{\bf 0.25}$\\
\hline
\end{tabular}
\end{table}

\begin{figure}[!h]
\centering
  \subfigure[LCC]{
    \label{LCC-confMat}
      \begin{minipage}[t]{0.7 \textwidth}
      \flushleft
      \includegraphics[width=10cm]{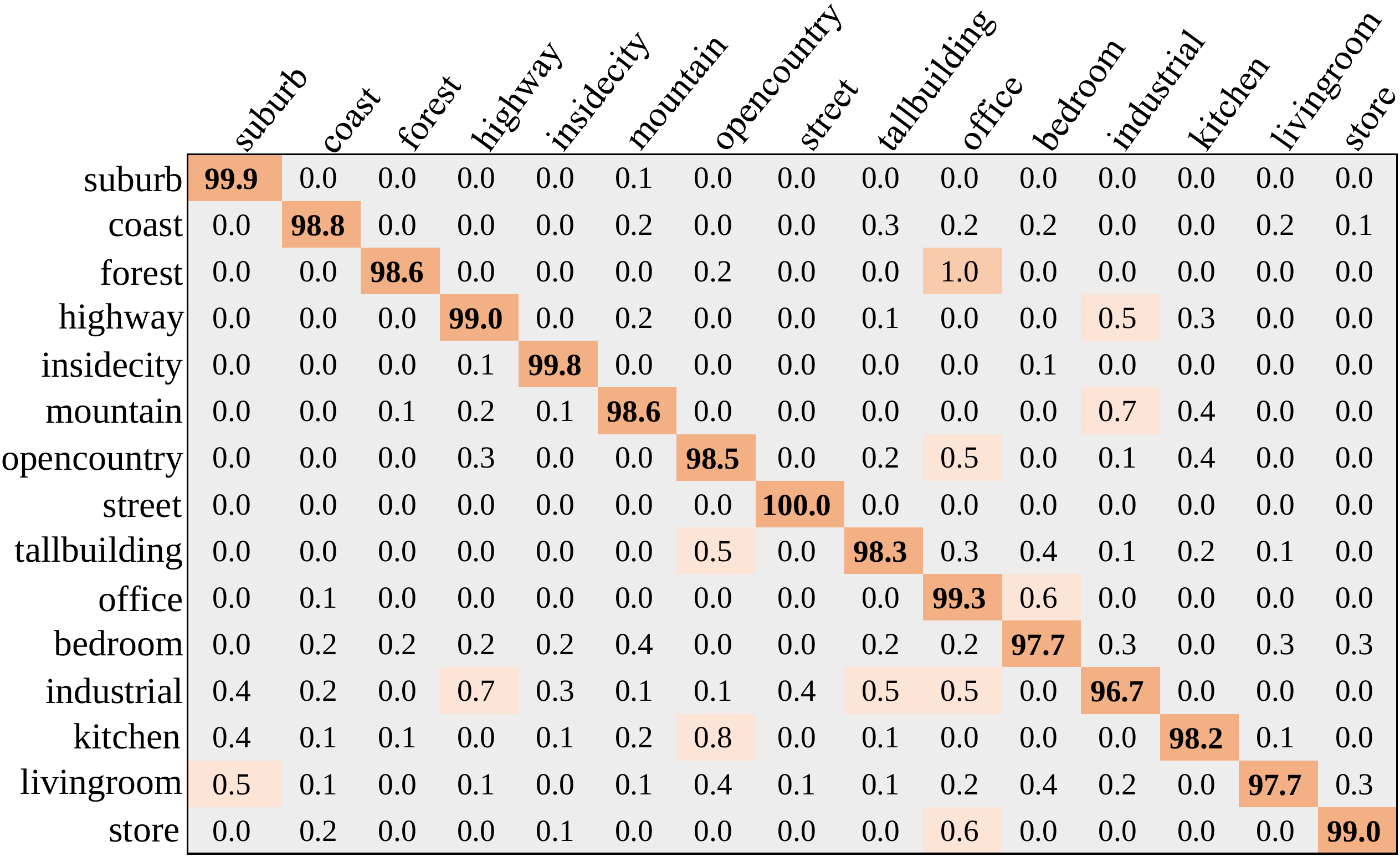}
      \end{minipage}
    }
    \\
  \subfigure[GCC]{
    \label{GCC-confMat}
      \begin{minipage}[t]{0.7 \textwidth}
      \flushleft
      \includegraphics[width=10cm]{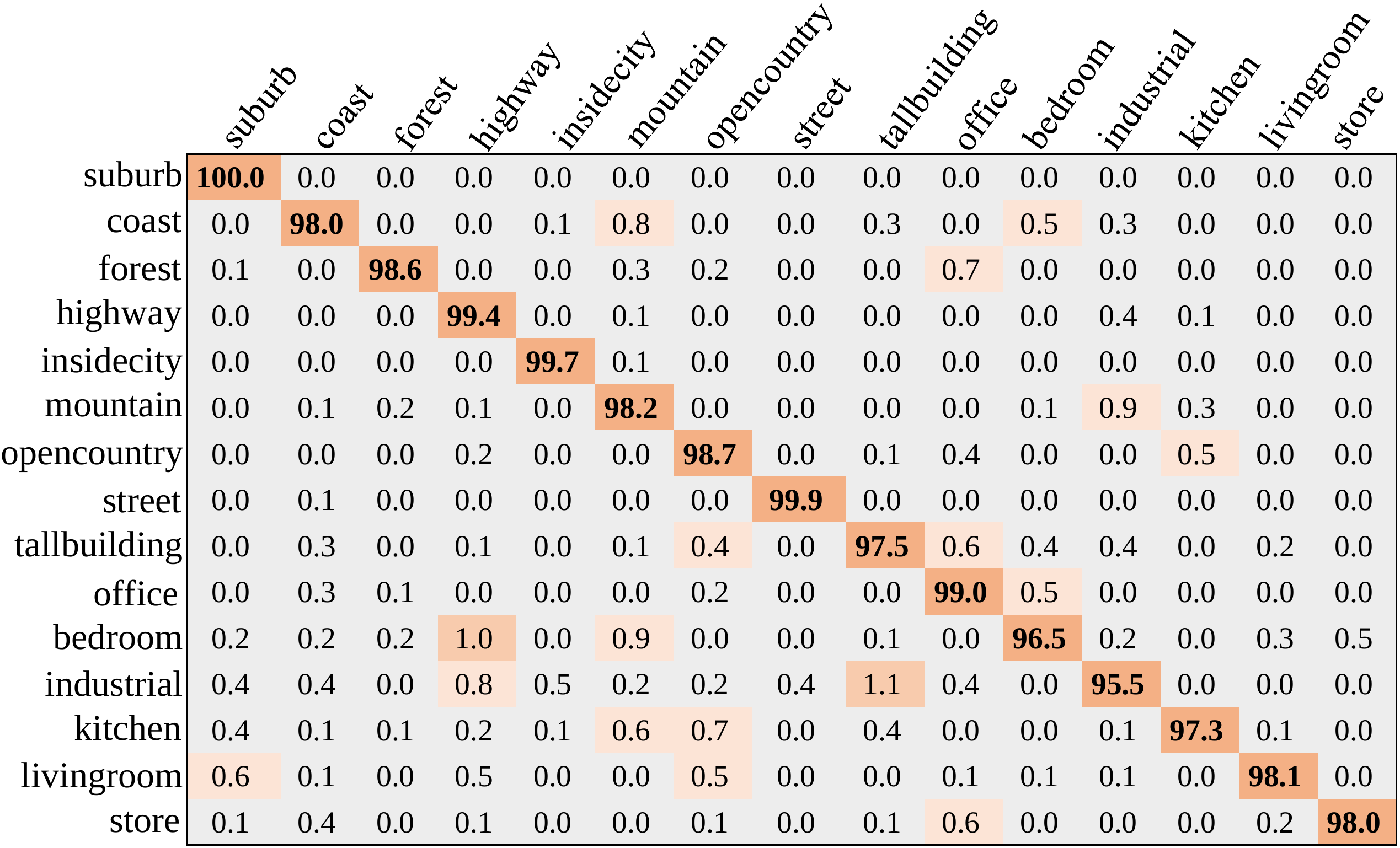}
      \end{minipage}
    }
  \caption{Confusion matrices on the 15 scene categories with the dictionary size $H=450$ using LCC and GCC, respectively.}
  \label{Scene15-fig}
\end{figure}

\subsection{DynTex++}
Dynamic textures are videos or sequences of moving scenes exhibiting certain stationary properties.
The DynTex++ dataset proposed in \cite{B_Ghanem_Dynamic2010} is a challenging dynamic texture dataset,
which is composed of 36 categories of textures ranging from waves on beach to branches swaying in wind (examples shown in Figure \ref{Dyntex_samp}).
Furthermore, each category contains 100 sequences with a fixed size $50 \times 50 \times 50$. The 177-dimensional LBP-TOP histogram \cite{G_Zhao_LBPTOP_2007}
is extracted from each dynamic texture sequence for categorization.
One half of the samples are adopted for training and the remaining ones are used for test.
The dictionary size for our method is set as $36\times 30=1080$ without any shared atoms.
The parameters $\beta,\ \lambda$, and $\gamma$ are optimized as $2\times 10^{-3}$, $2\times 10^3$, and $1\times 10^{-1}$ respectively in this experiment.
In addition to SRC \cite{J_Wright_2009}, K-SVD \cite{K-SVD_M_2006}, \ D-KSVD \cite{DKSVD_2010},
Joint \cite{D_Pham_2008}, DLSI \cite{DLSI_I_Ramirez_2010}, LC-KSVD \cite{LCKSVD2013}, COPAR \cite{COPAR2014}, we compare another developed dictionary based approaches Grassmann manifolds based method \cite{M_Harandi_2015}
and other state-of-the-art methods like \cite{ B_Ghanem_Dynamic2010, G_Zhao_LBPTOP_2007,GGDA_2011}.
The results based on 10 times random training/test splits are listed in Table \ref{Dyntex_comparison}.
\begin{figure}[t]
\begin{center}
\includegraphics[width=10cm]{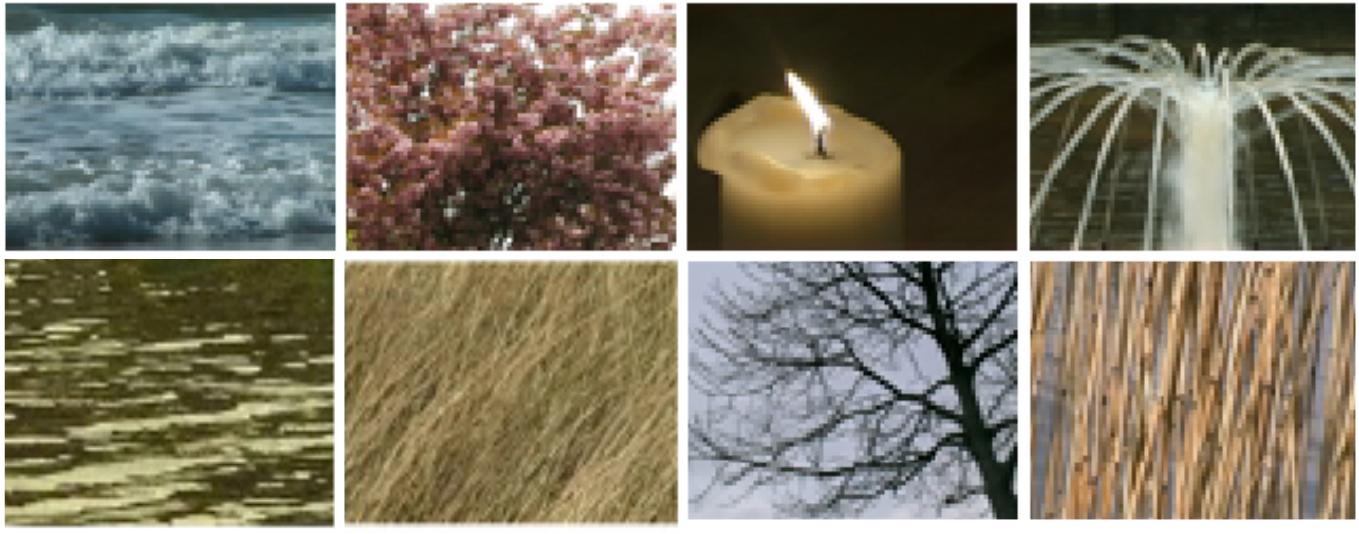}
\caption{Examples from different categories in Dyntex++ dataset.}
\label{Dyntex_samp}
\end{center}
\vspace{-1em}
\end{figure}

\begin{table}[t]
\caption{Accuracy on Dyntex++ dataset}
\label{Dyntex_comparison}
\centering
\begin{tabular}{|c|c|c|c|}
\hline
Method&Accuracy\,(\%)&Method&Accuracy\,(\%)\\[0.2ex]
\hline\hline
Ghanem \cite{B_Ghanem_Dynamic2010}   &63.70                           &Zhao \cite{G_Zhao_LBPTOP_2007}              &89.80\\
GGDA \cite{GGDA_2011}                &84.10                           &Xu \cite{Y_Xu_2011}                         &89.90\\
SVM \cite{SVM}                       &90.85$\,\pm\,0.28$              &COPAR \cite{COPAR2014}                      &94.32$\,\pm\,0.17$\\
SRC \cite{J_Wright_2009}             &88.53                           &D-KSVD \cite{DKSVD_2010}                    &89.27\\
DLSI \cite{DLSI_I_Ramirez_2010}      &91.56$\,\pm\,1.22$              &kgSC-dic \cite{M_Harandi_2015}              &92.80\\
K-SVD \cite{K-SVD_M_2006}            &89.31                           &kgLC-dic \cite{M_Harandi_2015}              &93.20\\
Joint \cite{D_Pham_2008}             &89.40                           &MCDL \cite{Y_Quan_MultiClassifier2016}      &90.35\\
LC-KSVD \cite{LCKSVD2013}            &89.67                           &{\bf ours\,(LCC)}                            &{\bf 95.72}$\,\pm\,{\bf 0.50}$\\
\hline
\end{tabular}
\vspace{-1em}
\end{table}

It can be seen that our method with LCC outperforms out all the competing methods with the highest accuracy.
Among them, kgSC-dic \cite{M_Harandi_2015} and kgLC-dic \cite{M_Harandi_2015} based on sparse and locality-constrained coding respectively are
the current state-of-the-art methods for this dataset, their results are obviously inferior to ours by above $2.5\%$.
Besides both kernel and Grassmann manifold strategies are applied in these two methods, and then they are more complicated than our algorithm.
The developed dictionary methods such as DLSI \cite{DLSI_I_Ramirez_2010} and COPAR \cite{COPAR2014} also achieve impressive recognition rates.
However, according to the computation efficiency comparison list in Table \ref{Dyntex_comparison},
they are more time-consuming for the dictionary learning and especially for categorization than ours.

\begin{table}[t]
\caption{computation efficiency on DynTex++ database}
\label{Dyntex-computation}
\centering
\begin{tabular}{|c|c|c|}
\hline
Method&Training time\,(s)&Time per test samp.\,(ms)\\
\hline\hline
DLSI \cite{DLSI_I_Ramirez_2010}     &35.64  &66.92\\
COPAR \cite{COPAR2014}              &520.22 &64.17\\
\hline
{\bf ours\,(LCC)}                   &{\bf 26.92}  &{\bf 0.17}\\
\hline
\end{tabular}
\end{table}

\subsection{UCF50}
UCF50 \cite{UCF50} is an sport action recognition dataset with $50$ action categories and over $6600$ videos, composed of realistic videos collected from YouTube,
including baseball pitch, basketball shooting, bench press, biking, billiards shot, breaststroke, etc, shown in Figure \ref{UCF50-demo}.
Due to large variations in camera motion, object appearance, pose, object scale, viewpoint and so on, this action dataset is very of challenge.
The action bank features \cite{Sadanand-2012} are adopted in our work for their superior performance in action recognition and
the original feature dimensionality is further reduced to 6000 from about 15000 by PCA for fast computation.
Our dictionary size is set to $75\times 50 =3750$ with 75 label-particular atoms for each class and no shared atoms.
The parameters $\beta,\ \lambda$, and $\gamma$ are optimized as $2\times 10^{-3}$, $2\times 10^3$, and $1\times 10^{-1}$ in this experiment, respectively.

Following the experimental setting in \cite{Sadanand-2012}, we evaluate our approach with five-fold group-wise cross validation scheme,
where given all the data has been divided into five folds, one fold is used for testing and the remaining four folds are applied for training.
We compare our proposed method with dictionary approaches including SRC \cite{J_Wright_2009}, D-KSVD \cite{DKSVD_2010}, DLSI \cite{DLSI_I_Ramirez_2010}, LC-KSVD \cite{LCKSVD2013}, COPAR \cite{COPAR2014}, FDDL \cite{FDDL_D_Yang2011}, and JDL \cite{JDL-2012},
and other state-of-the-art methods \cite{ Sadanand-2012, Oliva-2001, Wang-2009}.
In Table \ref{UCF50-comparison}, it's shown that our approach achieves a better result than all the other compared ones again, and outperforms the second best algorithm \cite{FDDL-2014} by
an evident improvement of $3.1\%$
\begin{figure}[!t]
\begin{center}
\includegraphics[width=10cm]{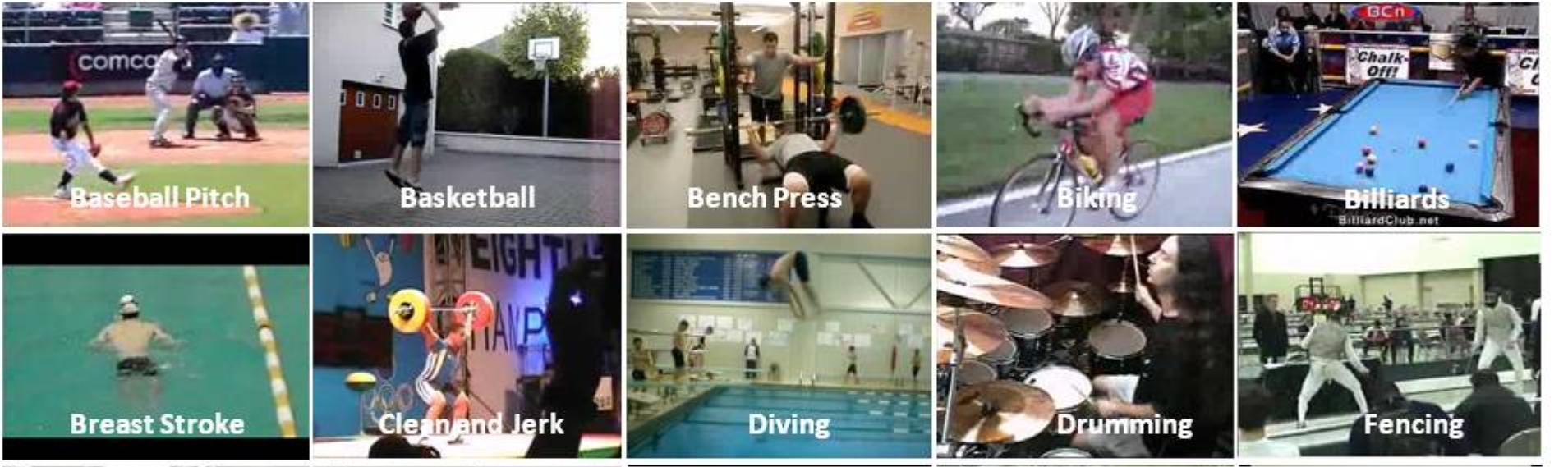}
\caption{Examples from different categories in UCF50 dataset.}
\label{UCF50-demo}
\end{center}
\end{figure}

\begin{table}[!t]
\caption{Accuracy on UCF50 dataset}
\label{UCF50-comparison}
\centering
\begin{tabular}{|c|c|c|c|}
\hline
Method&Accuracy\,(\%)&Method&Accuracy\,(\%)\\[0.2ex]
\hline\hline
Olive \cite{Oliva-2001}        &38.80            &LC-KSVD \cite{LCKSVD2013}           &53.60\\
Wang \cite{Wang-2009}          &47.90            &FDDL \cite{FDDL-2014}               &61.10\\
SVM  \cite{Sadanand-2012}      &57.90            &COPAR \cite{COPAR2014}              &52.50\\
SRC \cite{J_Wright_2009}       &59.60            &K-SVD \cite{K-SVD_M_2006}           &47.70$\,\pm\,1.54$\\
JDL \cite{JDL-2012}            &53.50            &DLSI \cite{DLSI_I_Ramirez_2010}     &55.91$\,\pm\,2.45$\\
D-KSVD \cite{DKSVD_2010}       &38.60            &{\bf ours\,(GCC)}                   &{\bf 64.21}${\bf \,\pm\,2.21}$\\
\hline
\end{tabular}
\end{table}

\section{Conclusions}
In this paper, considering a structured dictionary consisting of label-particular atoms and shared atoms,
we propose cross-label suppression dictionary learning with group regularization to leverage the discriminative power.
We don't resort to employing time-consuming $\ell_0$-norm or $\ell_1$-norm for regularizing representations.
As a result, the learning process and coding for classification become very fast.
Moreover, a wealth of experiments demonstrate our proposed approach can obtain promising classification results for extensive tasks,
including face recognition, object classification, scene categorization, texture recognition and action categorization, and outperform SVM \cite{SVM} and recently proposed dictionary methods,
like SRC \cite{J_Wright_2009}, D-KSVD \cite{DKSVD_2010}, DLSI \cite {DLSI_I_Ramirez_2010}, LC-KSVD \cite{LCKSVD2013}, COPAR \cite{COPAR2014}, etc.

Due to kernel and manifold techniques can address the nonlinear problem better than direct linear reconstruction for signals, in future,
we can incorporate these strategies to extend our model for further improvement.
Besides, a classifier can be jointly trained with the dictionary, and the effectiveness has been demonstrated in \cite{Mairal-SDL2009}, \cite{DKSVD_2010}, \cite{LCKSVD2013}.
Therefore, we will study how to train an effective classifier during the dictionary learning, to attain a better performance on recognition accuracy and speed.

\clearpage

\end{document}